\pdfoutput=1

\documentclass[11pt,dvipsnames]{article}

\usepackage[preprint]{acl}

\usepackage{times}
\usepackage{latexsym}

\usepackage[T1]{fontenc}

\usepackage[utf8]{inputenc}
\usepackage{subcaption}
\usepackage{CJKutf8}
\usepackage{microtype}
\usepackage{inconsolata}
\usepackage{amsmath}
\usepackage{amssymb}
\usepackage{xspace}
\usepackage{siunitx}
\usepackage{hyperref}
\usepackage{graphicx}
\usepackage{booktabs} 
\usepackage{enumitem}
\usepackage{pgfplots}
\usepackage{tikz}
\usepackage{listings}
\usepackage{colortbl}
\usepackage{xcolor}
\usepackage{multirow}
\usepackage{multicol}
\usepackage{algorithm}
\usepackage{algpseudocode}

\definecolor{questioncolor}{rgb}{0.3098, 0.5412, 0.4196}
\definecolor{conclusioncolor}{rgb}{0.6588, 0.4235, 0.4314}

\newcommand{\coloredquestiontt}[1]{\textcolor{questioncolor}{\textbf{\texttt{#1}}}}
\newcommand{\coloredconclusiontt}[1]{\textcolor{conclusioncolor}{\textbf{\texttt{#1}}}}
\title{Implicit Reasoning in Transformers is Reasoning through Shortcuts}

\author{Tianhe Lin\textsuperscript{\rm $\heartsuit$},
 Jian Xie\textsuperscript{\rm $\spadesuit$},
 Siyu Yuan\textsuperscript{\rm $\heartsuit$},
 \textbf{Deqing Yang}\textsuperscript{\rm $\heartsuit$}\thanks{Corresponding author.}\\
\textsuperscript{\rm $\heartsuit$}School of Data Science, Fudan University\\
\textsuperscript{\rm $\spadesuit$}College of Computer Science and Artificial Intelligence, Fudan University \\
\texttt{\{thlin20,yangdeqing\}@fudan.edu.cn} \\ 
\texttt{\{jianxie22, syyuan21\}@m.fudan.edu.cn}
}

\begin{document}
\maketitle
\begin{abstract}
Test-time compute is emerging as a new paradigm for enhancing language models' complex multi-step reasoning capabilities, as demonstrated by the success of OpenAI's o1 and o3, as well as DeepSeek's R1.
Compared to explicit reasoning in test-time compute, implicit reasoning is more inference-efficient, requiring fewer generated tokens.
However, \textit{\textbf{why does the advanced reasoning capability fail to emerge in the implicit reasoning style?}}
In this work, we train GPT-2 from scratch on a curated multi-step mathematical reasoning dataset and conduct analytical experiments to investigate how language models perform implicit reasoning in multi-step tasks.
Our findings reveal:
1) Language models can perform step-by-step reasoning and achieve high accuracy in both in-domain and out-of-domain tests via implicit reasoning. However, this capability only emerges when trained on fixed-pattern data.
2) Conversely, implicit reasoning abilities emerging from training on unfixed-pattern data tend to overfit a specific pattern and fail to generalize further. Notably, this limitation is also observed in state-of-the-art large language models.
These findings suggest that language models acquire implicit reasoning through shortcut learning, enabling strong performance on tasks with similar patterns while lacking generalization.
Resources are available on the \href{https://github.com/TianheL/LM-Implicit-Reasoning}{GitHub}.

\end{abstract}

\section{Introduction}
\label{sec:intro}
\label{intro}
Chain-of-Thought (CoT; \citet{NEURIPS2022_9d560961}) has sparked the development of explicit reasoning in large language models (LLMs). 
The subsequent rise of large reasoning models~\citep{o1,geminithinking, deepseekai2025deepseekr1incentivizingreasoningcapability} based on long CoT demonstrates impressive capabilities across various tasks~\citep{rein2023gpqagraduatelevelgoogleproofqa,aime,jimenez2024swebench}. 
Recent works have shown that such reasoning capabilities can even be distilled into smaller models~\citep{deepseekai2025deepseekr1incentivizingreasoningcapability}.\footnote{In this paper, ``smaller'' is relative to super large LMs like Deepseek R1, which has 671B parameters.}
Different from explicit reasoning, implicit reasoning offers greater inference efficiency by relying on fewer tokens to generate an answer~\cite{deng2023implicitchainthoughtreasoning}. 
Yet, it falls short of the performance achieved by explicit reasoning~\cite{deng2024explicitcotimplicitcot,allenzhu2024physicslanguagemodels32}.
\textit{Why can't implicit reasoning develop advanced reasoning capabilities?}

While recent advances in mechanistic interpretability have aimed to demystify the implicit reasoning processes of language models (LMs), most studies are limited to single-step reasoning~\cite{NEURIPS2022_6f1d43d5, wang2023interpretability, nanda2023progress}, which does not meet the expectation for handling complex reasoning tasks, such as advanced mathematical problems.
Meanwhile, for multi-step implicit reasoning, previous work primarily focuses on reasoning over factual knowledge~\cite{yang-etal-2024-large-language-models,biran-etal-2024-hopping}, which may be hindered by issues such as inflated reasoning performance due to memorizing entity co-occurrences in the pre-training data~\cite{elazar2023measuringcausaleffectsdata, kang-choi-2023-impact, ju-etal-2024-investigating}.

In this paper, to minimize the impact of memorization and investigate the underlying reasoning mechanisms, we explore implicit reasoning through the lens of mathematical problems.
Mathematical reasoning primarily depends on arithmetic operations that follow strict logical rules, which require algebraic manipulation based on specific operators and operands rather than recalling pre-trained knowledge like entity relationships.
Given that the strength of explicit reasoning stems from stepwise rationales, the first question we seek to address is
\coloredquestiontt{RQ1: Can language models perform stepwise reasoning internally?}
To investigate this, we train GPT-2 from scratch on our synthetic multi-step dataset composed of sequential mathematical operations, which means premises are arranged in the same order as they appear in the actual step-by-step calculation process.
The experimental results and activation patching~\cite{NEURIPS2020_92650b2e, NEURIPS2022_6f1d43d5} plots show that \coloredconclusiontt{LMs can fully learn to do stepwise reasoning internally and generalize to problems with more steps, provided they are trained on data where all premises are presented sequentially.}

\begin{figure}[t]
    \centering
    \includegraphics[width=\linewidth]{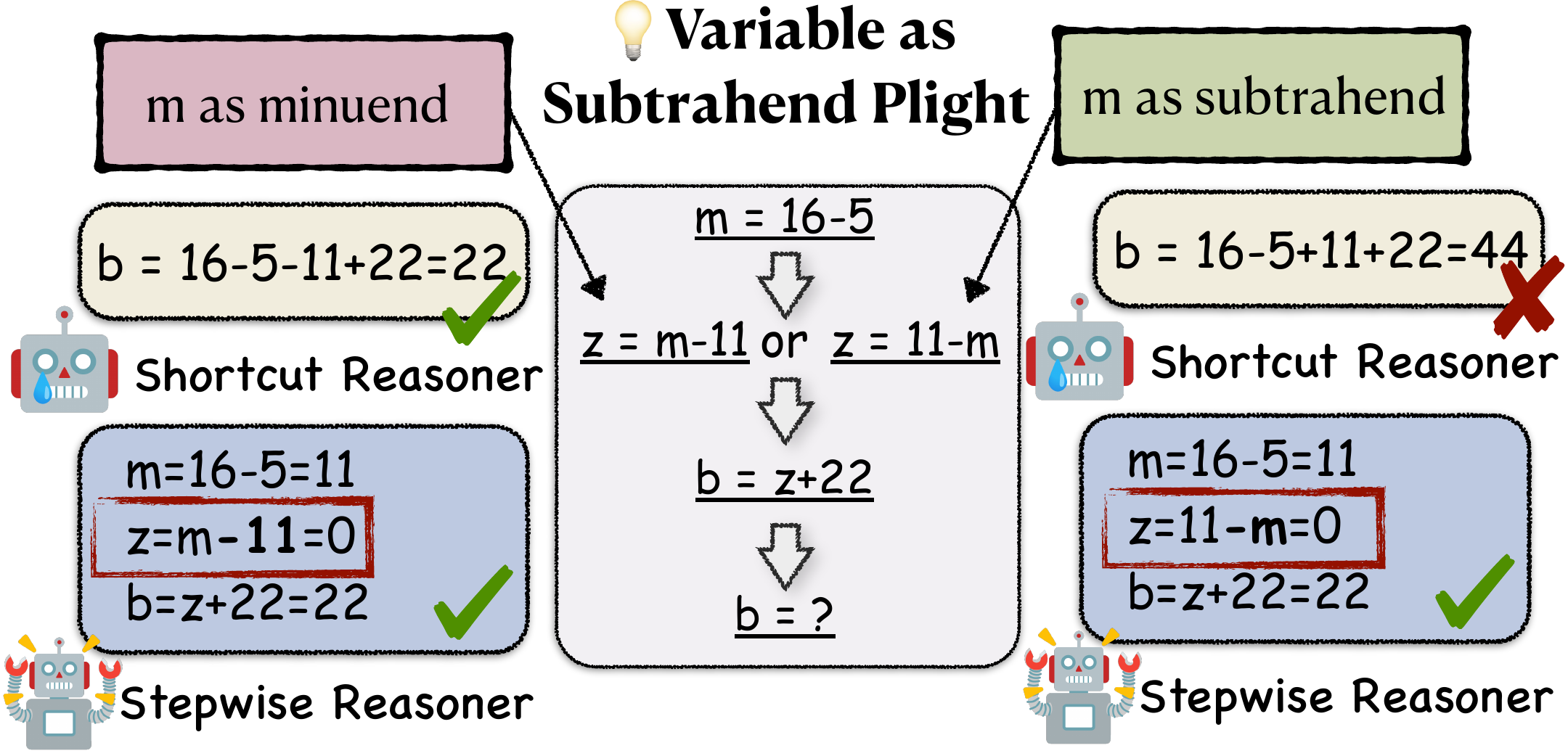}
     \vspace{-1.5em}
    \caption{A failure of generalization in language models trained on data with unfixed patterns, namely ``Variable as Subtrahend Plight''. 
    When trained on unfixed premise order, the model learns a reasoning shortcut that benefits from addition commutativity. This shortcut enables the model to perform implicit reasoning by chaining numbers, which fails when variables are subtrahends.
    }
     \vspace{-1.5em}
    \label{fig:intro_fig}
\end{figure}

However, the premises are not always presented sequentially in real-world reasoning tasks, requiring LMs to organize the information internally.
Therefore, based on the findings from RQ1, this paper seeks to answer a more general research question, \coloredquestiontt{RQ2: How do language models think internally if the premise order is not fixed?}
In contrast to the accuracy saturation in RQ1, accuracy drops significantly when the premise order is unfixed.
We conduct further analysis and find \coloredconclusiontt{LMs fail to learn stepwise implicit reasoning when the premise order is not fixed, struggling with ``Variable as Subtrahend Plight''.}
Specifically, as shown in Figure~\ref{fig:intro_fig}, models trained on an unfixed premise order overfit to an easy pattern in the data, relying on a shortcut that benefits from addition commutativity. 
This shortcut allows the model to solve the problem by directly chaining numbers, while the presence of variables in the subtrahend position disrupts this shortcut.
Additional mechanistic analysis validates our hypothesis.

Previous work demonstrated that even current state-of-the-art (SoTA) LLMs also struggle with implicit reasoning~\citep{yu2024llmsreallythinkstepbystep}.
Based on our previous findings, we aim to investigate \coloredquestiontt{RQ3: How do LLMs perform multi-step implicit reasoning?} 
We find \coloredconclusiontt{``Variable as Subtrahend Plight'' also persists in SoTA LLMs, indicating that these models, trained on diverse unfixed premise corpora, are also relying on shortcuts for multi-step implicit reasoning.}
This further validates the correctness and generalizability of our findings.

To summarize, in this paper, we investigate the internal mechanisms of implicit reasoning in transformers and uncover why the advanced reasoning capabilities observed in explicit reasoning do not emerge in implicit reasoning.
While we reveal that current LMs primarily rely on shortcuts for implicit reasoning, a silver lining is that a stepwise reasoning pattern could indeed emerge through training.
Such a pattern underpins the advanced reasoning capabilities of LMs, and we envision that future advanced strategies could help form this pattern.

\section{Related Work}
\label{sec:related}
\subsection{Mechanistic Interpretability of Language Models}
Mechanistic interpretability (MI) aims to uncover and explain the internal workings of models.
The research of mechanistic interpretability in language models primarily focuses on three key areas: \textit{features} within model representations~\cite{nostalgebraist2020, gurnee2023finding,zhou2024pretrained}, \textit{circuits} connecting these features~\cite{wang2023interpretability,NEURIPS2023_efbba771,prakash2024finetuning}, and \textit{universality} across diverse models and tasks~\cite{chughtai2023toy,gurnee2024universal}.

Mathematical tasks, due to their significance in representing the reasoning capabilities of language models, have been widely studied in MI~\cite{NEURIPS2023_efbba771,kudo2024thinktotalktalktothinkllmscome,zhou2024pretrained}.
However, most of the existing studies~\cite{stolfo-etal-2023-mechanistic,yu-ananiadou-2024-interpreting,pmlr-v235-zhang24bk,chen2024stateshiddenhiddenstates} focus on single-step mathematical reasoning.
How LMs perform multi-step mathematical reasoning implicitly remains poorly understood.
To bridge this gap, we employ activation patching~\cite{NEURIPS2020_92650b2e} to track the information flow and reverse-engineer the behaviors of LMs in multi-step arithmetic computations.

\subsection{Multi-step Implicit Reasoning}
As opposed to explicit reasoning, implicit reasoning is performed in the hidden states instead of extra tokens. 
Previous studies typically investigate implicit reasoning in two domains: factual reasoning~\cite{wang2024grokking,yang-etal-2024-large-language-models,yang2024largelanguagemodelsperform,biran-etal-2024-hopping} and mathematical reasoning~\cite{stolfo-etal-2023-mechanistic,nanda2023progress,deng2024explicitcotimplicitcot}.
However, progress in reasoning over factual knowledge risks being inflated by entity co-occurrence learned from pre-training data~\cite{elazar2023measuringcausaleffectsdata, kang-choi-2023-impact, ju-etal-2024-investigating}.
While mathematical reasoning is less susceptible to this issue due to the variability of operands and operators, LMs may rely on shortcuts or shallow heuristics to predict the results~\cite{liu2023transformers,nikankin2025arithmetic,xie2024revealing}, which are often overlooked in studies on multi-step implicit reasoning.
In our study, we scrutinize the impact of shortcuts and represent the internal mechanisms driving the observed phenomenon to the investigation of the multi-step implicit mathematical reasoning abilities in Transformer-based LMs.

\section{General Setup}
\label{sec:general_setup}
\paragraph{Task.} 
Focusing on reasoning capability rather than other factors (e.g., factual knowledge memorization), we use mathematical problems as a lens. 
To further minimize the impact of natural language complexity, we shift our focus to mathematical formulas rather than problem statements in natural language. 
Specifically, we construct a synthetic dataset of multi-step sequential modular addition and subtraction as our testbed for analysis.
As shown in Figure~\ref{fig:intro_fig}, except for the first step, each step of the computation involves a variable from the previous step, a number (we name it operand later), and an operator (i.e., ``$+$'' or ``$-$'').
Following \citet{ye2024physicslanguagemodels21}, we consider using arithmetics mod23 to avoid numbers being split into multiple tokens and prevent errors from large number calculations, thereby focusing on reasoning itself rather than calculation. 

\paragraph{Data.} 
For training data, we generate different multi-step calculation templates for questions at each length (ranging from 1 to 5 steps) and then randomly use $K$ different groups of variable names to instantiate each template.\footnote{$K = 2$ in this paper. Please refer to Appendix~\ref{appendix:choice-of-k} for more details about the data generation process.} 
To prevent LMs from memorizing intermediate results from the training set rather than performing actual reasoning to solve the math problems in our test set, we filter out all templates with preceding calculations, apart from the first step, that overlap with the templates of the training set during the test set generation process.
For example, if ``f=1+2,s=3-f,s>>?'' appears in the training set, then ``a=1+2,b=3-a,c=b+5,c>>?'' is not allowed to appear in the test set because the first two steps of the former are the same as the latter regardless of variable names.

We evaluate both in-distribution (ID) and out-of-distribution (OOD) performance, which are distinguished by the maximum reasoning steps of the training set, with ID not exceeding the maximum steps of the training set (i.e., 5-step) and OOD being one or two steps more than the maximum steps of the training set (i.e., 6-step or 7-step).
ID generalization aims to evaluate whether the model learns the latent rules of the training set, while OOD generalization is designed to assess whether the model genuinely acquires some reasoning skills.

\paragraph{Model \& Optimization.} 
Following \citet{ye2024physicslanguagemodels21}, we use a standard 12-layer GPT-2 model~\cite{radford2019language} and replace its positional embeddings with rotary embeddings (RoPE)~\cite{10.1016/j.neucom.2023.127063} to enable the model to learn length generalization (i.e., to generalize its ability to solve more steps in multi-step reasoning tasks than those seen during training).
We use AdamW~\citep{loshchilov2018decoupled} with learning rate $10^{-4}$, batch size \num{1600}, weight decay \num{0.1} and \num{2000} warm-up steps.

\paragraph{Activation Patching.} 
Activation patching~\cite{NEURIPS2020_92650b2e,NEURIPS2022_6f1d43d5} is a strategy for identifying the important modules that causally affect the output by intervening on their latent activations.
Specifically, if a module is important, the alteration of its activation will significantly affect the model's output, whereas an unimportant one will have little to no impact.
Typically, the method needs two inputs, an original one (e.g., ``a=\underline{1}+4,d=5-a,c=1+d,c>>?'') and another with a slight difference (e.g., ``a=\underline{6}+4,d=5-a,c=1+d,c>>?''), and three forward passes: 
The \textbf{clean run} and the \textbf{corrupted run} take the above two inputs separately and cache activations of the model's components, such as attention or MLP outputs. 
In the \textbf{patched run}, we run the model on the original input but replace the specific activation with the cached activation from the corrupted run.
Following previous work~\cite{pmlr-v235-zhang24bk}, we measure the changes in the output logits of the ground truth tokens.
Then, we compute the patching effect ($\operatorname{PE}$) as:
\begin{equation}
\label{eq:pe}
\operatorname{PE}=\frac{\operatorname{Logit}_\mathrm{cl}(r) - \operatorname{Logit}_\mathrm{pt}(r)}{\operatorname{Logit}_\mathrm{cl}(r)},
\end{equation}
where $r$ is the correct answer of the original input and $\mathrm{cl}$, $\mathrm{pt}$ denote the clean and patched run separately.
The experiments are conducted on 100 randomly selected samples.

We iterate activation patching over a set of activations and compare how much they affect the final output, which allows us to localize which activation matters and ultimately reverse-engineer the underlying circuit.
In practice, we utilize sliding window patching~\cite{hase2023does} with window size $2\times2$, where at each token position, the representations of the $2\times2$ region formed by the current layer and the next layer, along with the current token and the next token, are substituted by the cached activations from the corrupted run.\footnote{We study the choice of metrics in Appendix~\ref{appendix:choice-of-metric} and window sizes in Appendix~\ref{appendix:window-size}.}

\section{Can Language Models Perform Stepwise Reasoning Internally?}
\label{sec:rq1}
Previous work found that smaller LMs ($\sim$7B) can hardly do multi-step mathematical reasoning correctly without CoT, while a 70B level model can only achieve an accuracy of about 50\% in 4-hop reasoning~\cite{yu2024llmsreallythinkstepbystep}. 
Since previous work demonstrated that externalizing reasoning step by step enhances performance in mathematical tasks~\citep{NEURIPS2022_9d560961}, a question is: \textit{does the poor performance of implicit reasoning arise from the inability to employ this step-by-step reasoning style?}
We begin our investigation by training our GPT-2 model on the synthetic dataset to learn implicit reasoning.

\subsection{Results}
\begin{figure}[t]
    \centering
    \includegraphics[width=0.97\linewidth]{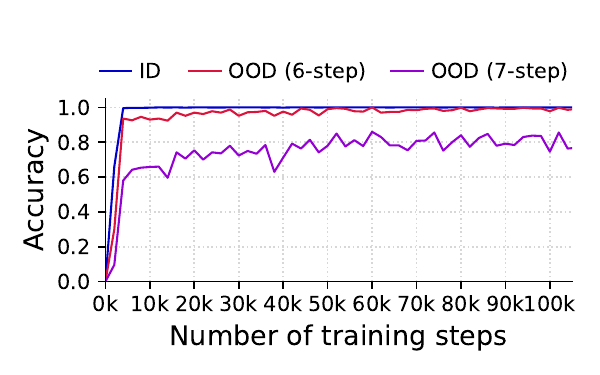}
    \vspace{-0.5em}
    \caption{Test accuracy during the training stage. We find that Transformers are able to learn to reason implicitly and generalize well to those that require longer reasoning steps.} 
    \vspace{-1em}
    \label{fig:rope-seq-train-acc}
\end{figure}

\paragraph{Language models are able to perform implicit mathematical reasoning with near-complete accuracy when trained.}
We first analyze whether our model is capable of solving multi-step implicit mathematical reasoning.
Figure~\ref{fig:rope-seq-train-acc} shows the model's accuracy on both the ID and OOD test data throughout the optimization.
The model not only achieves 100\% accuracy on implicit reasoning tasks from the same distribution (ID set) but also \textbf{\emph{generalizes effectively to tasks requiring longer reasoning steps in the OOD set}}.
To be specific, the model achieves 99\% accuracy in tasks that require an additional step of reasoning and nearly 90\% accuracy in tasks that require two more.
This implies that the model truly learns some implicit reasoning skills rather than simply memorizing answers, since our model has never seen any training example of the same length as in the test time.

\subsection{The Working Mechanism of Model}
\paragraph{Setting.}
To investigate whether the language model is based on understanding (i.e., gathering all the information together first and then computing) or reasoning step by step, we use activation patching to reveal the model's internal thought process.
Two different experiment setups are used to reveal how the information is transmitted and what information is transmitted separately.

\noindent
$\bullet$ \textbf{\textit{Tracing the information flow.}}
To gain insights into the working mechanisms of the model, we first need to know the path through which the token's information is transmitted to the output, i.e., how the information of a specific token affects the output.
To this end, we change only one operand or operator in the original input and identify the activations that have an influence on the final output by replacing activations. 

\noindent
$\bullet$ \textbf{\textit{Tracking result-related information.}}
The first setting explains how information is transmitted to the output, yet what information is transmitted is still unclear. 
Therefore, we formulate a variant of the first setting to track the information related to intermediate results (i.e., the value of an intermediate variable). 
Specifically, we modify a set of operands and compare the differences in patching effects when the intermediate results are either identical or distinct.
For example, if we aim to track the related information of ``d'' in ``a=4+\underline{6},\textbf{d}=a+\underline{5},c=1+d,c>>?'', 
We need to modify the operands while keeping the value of ``d'' fixed at 15 (e.g., ``a=4+\underline{1}, \textbf{d}=a+\underline{10},c=1+d,c>>?'') and compare it with a case where the result changes (e.g., ``a=4+\underline{1},\textbf{d}=a+\underline{4},c=1+d,c>>?”, where d=9).
If the model is performing step-by-step reasoning, the patching effect will be more pronounced in the first two steps due to the change in ``a'' and will then diminish from the third step onward in the fixed-result setting, as the subsequent results remain unchanged.
\paragraph{Language models are able to do step-by-step reasoning internally.}
\begin{figure}[t]
    \centering
    \includegraphics[width=0.9\linewidth]{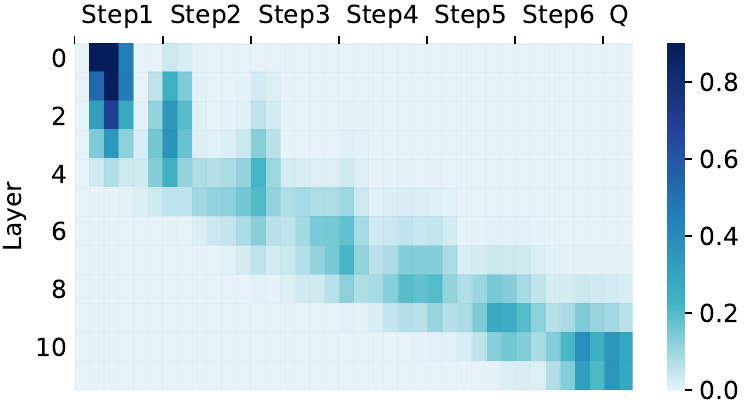}
    \caption{Activation patching on residual stream across layers and token positions when changing the first number in the problems. All the premise orders are forward.
    }
    \vspace{-0.5em}
    \label{fig:rq1-step-by-step}
\end{figure}

To trace the information flow, we first examine the residual stream patching plot by only altering one operand.\footnote{We show the information flow related to operators in Appendix~\ref{appendix:operator-related-info-flow}.}
The patching effects across layers and positions are shown in Figure~\ref{fig:rq1-step-by-step}.
We observe that a significant portion of the patching effects concentrate at the end of each step and exhibit a clear trend of gradually propagating along a diagonal line. 
This pattern forms the foundation of step-by-step reasoning, which implies that each intermediate result builds upon the last. 

\begin{figure*}[t]
    \centering
    \begin{subfigure}[b]{0.32\linewidth}
        \centering
        \includegraphics[width=5.2cm, keepaspectratio]{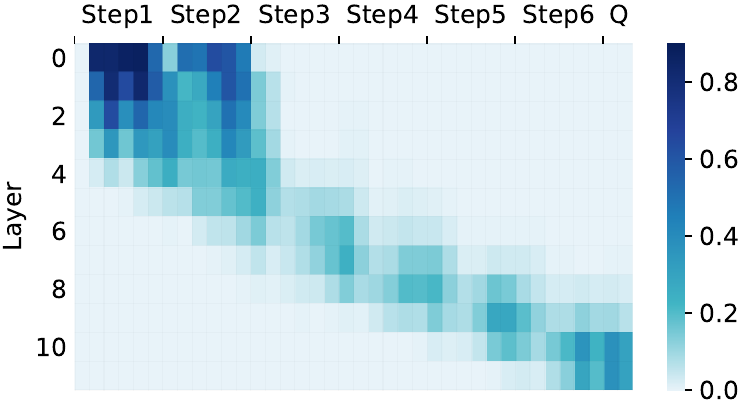}
        \caption{Residual Stream}
        \label{fig:hidden_diff}
    \end{subfigure}
    \hfill
    \begin{subfigure}[b]{0.32\linewidth}
        \centering
        \includegraphics[width=5.2cm, keepaspectratio]{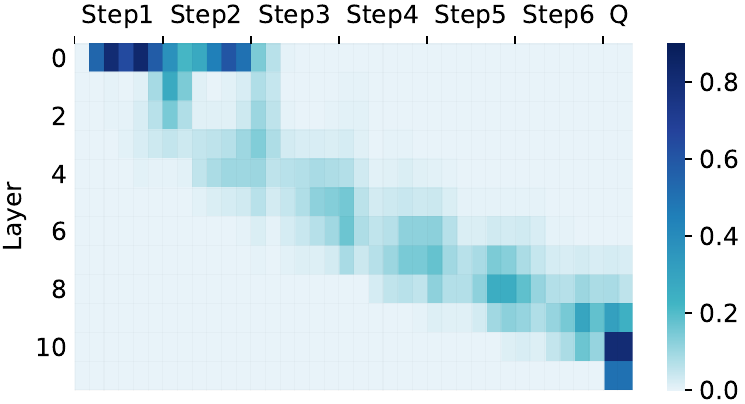}
        \caption{Attention}
        \label{fig:attn_diff}
    \end{subfigure}
    \hfill
    \begin{subfigure}[b]{0.32\linewidth}
        \centering
        \includegraphics[width=5.2cm, keepaspectratio]{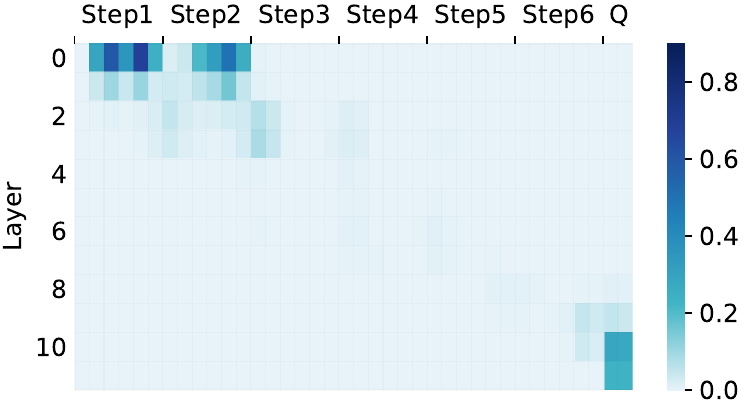}
        \caption{MLP}
        \label{fig:mlp_diff}
    \end{subfigure}
    \begin{subfigure}[b]{0.32\linewidth}
        \centering
        \includegraphics[width=5.2cm, keepaspectratio]{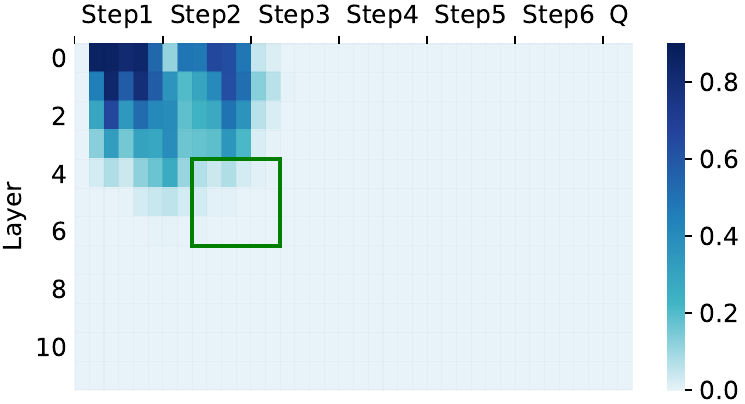}
        \caption{Residual Stream}
        \label{fig:hidden_same}
    \end{subfigure}
    \hfill
    \begin{subfigure}[b]{0.32\linewidth}
        \centering
        \includegraphics[width=5.2cm, keepaspectratio]{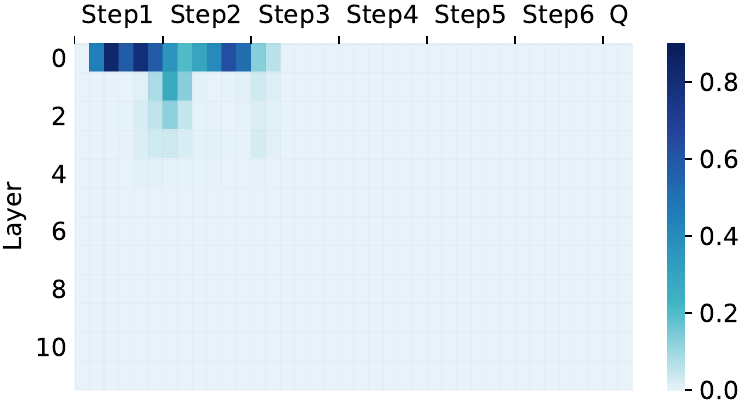}
        \caption{Attention}
        \label{fig:attn_same}
    \end{subfigure}
    \hfill
    \begin{subfigure}[b]{0.32\linewidth}
        \centering
        \includegraphics[width=5.2cm, keepaspectratio]{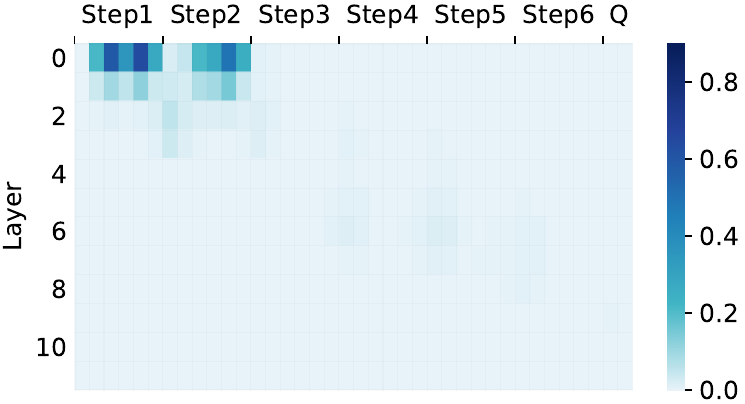}
        \caption{MLP}
        \label{fig:mlp_same}
    \end{subfigure}
    \caption{Patching effect of different components across layers and token positions. We change the numbers in the first two steps. The result of step 2 is \textbf{changed} in sub-figure (a)(b)(c), while the result is kept \textbf{unchanged} in sub-figure (d)(e)(f). A deeper color indicates the significance of activation at that position. We add a green rectangle in the figure to better illustrate the location where the patching effect first starts to diminish.} 
     \vspace{-0.95em}
    \label{fig:rq1-comp}
\end{figure*}

Based on the discovered information flow, we investigate the information behind these activations by tracking result-related information and adding constraints to ensure the results remain the same when changing the input.
By comparing the results of the result-varied setting (Figure~\ref{fig:hidden_diff}) and the result-fixed setting (Figure~\ref{fig:hidden_same}), we find:
The region between Step 2 and Step 3, where the impact diminishes (highlighted by the green box in Figure~\ref{fig:hidden_same}), aligns precisely with the segment between the second and third steps in the information flow (Figure~\ref{fig:rq1-step-by-step}). 
In the fixed-result setting, the substituted activations retain the same information, leading to a minor patching effect.
However, in the unfixed setting, the patching effect is more pronounced.
This provides evidence that this area stores information related to the intermediate results.

\begin{figure}[t]
    \centering
    \pgfplotsset{width=0.8\linewidth,height=0.45\linewidth,compat=1.18}
\begin{tikzpicture}
\begin{axis}[
    xmin=0.25, xmax=10.75,
    ymin=-0.1, ymax=1.1,
    xtick={1,2,3,4,5,6,7,8,9,10},
    ytick={0.0, 0.25, 0.5, 0.75, 1.0},
    ymajorgrids=true,
    xmajorgrids=true,
    grid style=dashed,
    xlabel={Attention Window Size},
    ylabel={Accuracy},
    x label style={at={(axis description cs:0.5,-0.225)},anchor=north},
    y label style={at={(axis description cs:-0.185,0.5)},anchor=south},
    legend style={nodes={scale=0.85}, legend columns=3,anchor=north,at={(0.48,-0.32)}}
]
\addplot[
    color=NavyBlue,
    mark=o,
    line width=0.5pt,
    mark size=2.6pt,
    error bars/.cd,
    y dir=both, y explicit,
    error bar style={line width=0.7pt, color=NavyBlue},
    error mark options={rotate=90, NavyBlue, mark size=2.5pt}
    ]
    coordinates {
    (1, 0.0)
    (2, 0.002)
    (3, 0.006)
    (4, 0.006)
    (5, 0.038)
    (6, 0.092)
    (7, 0.452)
    (8, 0.788)
    (9, 0.98)
    (10, 1)
    };
\end{axis}
\end{tikzpicture}
    \caption{Test accuracy under different attention window sizes on 5-step problems. A window size of $n$ means that a token can focus on itself and its preceding $n-1$ tokens.}
    \vspace{-0.5em}
    \label{fig:recover-rate}
\end{figure}
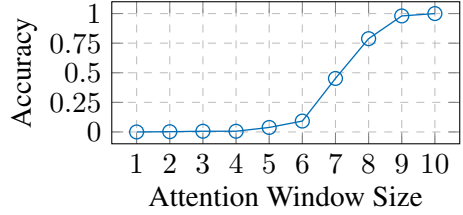

To further validate that the language model performs step-by-step reasoning by reusing intermediate results stored in a specific area from the last step, we conduct an additional experiment to examine its behavior when the information from previous steps is masked by varying the attention window size (see implementation details in Appendix~\ref{appendix:attn_window_size}).
Specifically, each step consists of 6 tokens, and when we scale up the attention window size to bigger than 6, the model is able to access the information stored from the previous step.
We present the model's accuracy under different attention window sizes in Figure~\ref{fig:recover-rate}. 
Our findings show that when the attention is restricted to the current step (i.e., window size = 6), the model completely loses its reasoning ability. 
However, once the attention is expanded to include the previous step's results, accuracy recovers rapidly. 
This supports the hypothesis that the model follows a step-by-step reasoning pattern, as evidenced from a different perspective.

To sum up, the model computes the result of each step once it concludes, and this information is then utilized by the subsequent step in the next layers, establishing a step-by-step computation pattern.

\paragraph{Attention mechanism propagates intermediate results, and MLP modules enhance features related to inputs and outputs.}
Intervening on hidden states only provides us with a glimpse of the information flow, but the roles of various components within the model remain unclear. 
By decomposing the causal effects of contributions of attention and MLP modules (Figure \ref{fig:attn_diff},\ref{fig:attn_same} and Figure \ref{fig:mlp_diff},\ref{fig:mlp_same}), we find a decisive role for attention modules in the middle layers and MLPs in early and final layers.
In conjunction with the findings on information flow, we infer that the attention layers are responsible for extracting the information needed in the current step and gradually transferring intermediate computational information to deeper layers. 
Therefore, a possible explanation of the model's behavior on this task is that the MLP modules enhance features of operators and operands in early layers, then the attention mechanism facilitates the step-by-step propagation of intermediate results, and finally, the MLP modules in the last layers enhance the probabilities of correct predictions.

\section{How Do Language Models Reason Internally When the Premise Order is Not Fixed?}
\label{sec:rq2}
Based on the above findings, we find that Transformers are able to perform step-by-step reasoning internally when the premise order is fixed.
However, in complex reasoning tasks, the premises are not always presented sequentially; they may appear in a random order, requiring LMs to organize the information internally.
\emph{Can language models still perform implicit reasoning step by step when the premises are shuffled?}

\paragraph{Setup.} 
For consistency, we continue to use the original data but randomly shuffle the order of premises\footnote{More details of data setups are included in Appendix~\ref{appendix:premise_pattern}.}, excluding the question. 
To assess the impact of premise order, we study three different orders, including forward, reverse, and random. 
Specifically, for the reverse order, we list the premises in reverse order; for the random order, we shuffle the premises randomly.

\subsection{Result}
\paragraph{Language models fail to learn implicit reasoning when the premise order is not fixed.}
In contrast to the high accuracy scores achieved by the model trained on fixed-order premises, Table~\ref{tab:rq2-acc-diff-step} shows that the model trained on shuffled premises fails to perform multi-step implicit reasoning correctly.
Specifically, as the number of steps increases, the model's accuracy gradually decreases, reaching only $\sim$40\% accuracy when five steps of reasoning are required, contrasting the saturated accuracy of the model trained on fixed premise order.

\begin{table}[t]
    \centering
    \small
    \begin{tabular}{lccccc}
    \toprule
        \textbf{Order} & \textbf{2-Step} & \textbf{3-Step} & \textbf{4-Step} & \textbf{5-Step} & \textbf{6-Step} \\ \midrule
        Forward & 1.00  & 0.87 & 0.57 & 0.43 & 0.23 \\ 
        Reverse & 1.00  & 0.81 & 0.51 & 0.38 & 0.19 \\ 
        Random & 1.00 & 0.83  & 0.53 & 0.37 & 0.23 \\ 
    \bottomrule
    \end{tabular}
\caption{
The accuracy of the model trained with unfixed premise order dataset on the original test set.
Each column represents problems with a specific number of steps, and each row represents a premise order used during testing.
}
\vspace{-0.75em}
\label{tab:rq2-acc-diff-step}
\end{table}

\paragraph{Language models struggle with ``Variable as Subtrahend Plight.''}
\begin{table}[t]
    \centering
    \small
    \begin{tabular}{lccccc}
    \toprule
    \multirow{2.5}{*}{\textbf{Order}} & \multicolumn{5}{c}{\resizebox{0.58\linewidth}{!}{\textbf{\textsc{\#Variable Being Subtrahend}}}}
    \\ \cmidrule(lr){2-6} 
         & 0 & 1 & 2 & 3 & 4 
         \\ 
        \midrule
\rowcolor[gray]{0.95} \multicolumn{6}{c}{\textit{3-Step Problems}} \\ \midrule
        Forward & 1.00 & 0.83 & 0.35 &  - &  -  \\ 
        Reverse & 1.00 & 0.73 & 0.15  & - & -    \\
        Random & 1.00 & 0.73 & 0.26  & - &  - \\ 
        \midrule
\rowcolor[gray]{0.95} \multicolumn{6}{c}{\textit{4-Step Problems}} \\ \midrule
        Forward & 1.00 & 0.33  & 0.08  & 0.12 &  -  \\ 
        Reverse & 1.00 & 0.15  & 0.08  & 0.10 & -   \\
        Random & 1.00 & 0.23 & 0.08 & 0.08  & -\\ 
        \midrule
\rowcolor[gray]{0.95} \multicolumn{6}{c}{\textit{5-Step Problems}} \\ \midrule
        Forward & 0.92 & 0.20 & 0.04  & 0.05  & 0.03  \\ 
        Reverse & 0.90  & 0.10  & 0.04 & 0.03 & 0.03   \\
        Random & 0.90  & 0.09  & 0.03  & 0.02 & 0.02 \\ 

    \bottomrule
    \end{tabular}
\caption{Accuracy of the model on questions with different numbers of variables being subtrahends. The accuracy is calculated on \num{100} instances. Since the first step involves an operation between two numbers, the maximum number of variables as subtrahends is one less than the total number of steps.}
\vspace{-1em}
\label{tab:rq2-acc-vas}
\end{table}
To explore how LMs perform implicit reasoning after being trained on an unfixed premise order, we conduct further analysis and find that the model is more prone to making mistakes when the premise contains multiple equations with a variable as the subtrahend.
Detailed statistics are provided in Table~\ref{tab:rq2-acc-vas} on how the model's accuracy varies with the number of variables being subtrahends for questions requiring three to five steps of reasoning.
As the number of variables being subtracted increases, the model's accuracy decreases drastically, which is consistent across different premise orders.
We term this phenomenon as ``Variable as Subtrahend Plight.''
When almost all the variables in the premise are subtrahends, the model almost fails to solve any of the problems correctly.
To rule out the possibility of a special case, we conduct experiments with increased data volume and with different models, yet the phenomenon remains consistent.
Please refer to Appendix~\ref{appendix:result-of-other-model} for more details.

To explore why models struggle with the ``Variable as Subtrahend Plight'', we revisit arithmetic expressions.
While addition benefits from commutativity (e.g., a+b=b+a), subtraction lacks this property, as swapping the minuend and subtrahend changes the outcome unless a=b.
This asymmetry creates challenges for models. 
For instance, in the sequence ``a=6+2,b=a-3,c=4+b'', the model might shortcut it as ``c=6+2-3+4'' (treating subtraction as addition).
However, when subtrahends are variables, such a shortcut fails. 
If ``b=3-a'', the model can no longer chain terms directly and must compute intermediate results in sequence.
As the number of variable subtrahends increases, the model faces greater difficulty in determining the correct order of operations, requiring rigorous step-by-step reasoning instead of relying on shortcuts.

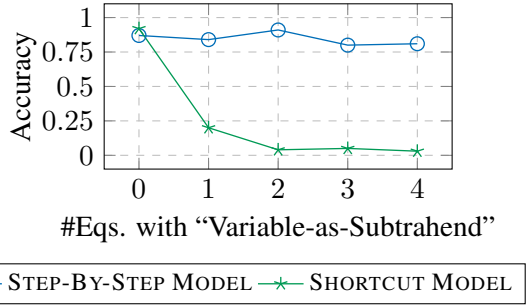
\begin{figure}[t]
    \centering
    \pgfplotsset{width=0.8\linewidth,height=0.49\linewidth,compat=1.18}
\begin{tikzpicture}
\begin{axis}[
    xmin=-0.5, xmax=4.5,
    ymin=-0.1, ymax=1.1,
    xtick={0, 1, 2, 3, 4, 5, 6},
    ytick={0.0, 0.25, 0.5, 0.75, 1.0},
    ymajorgrids=true,
    xmajorgrids=true,
    grid style=dashed,
    xlabel={\#Eqs. with ``Variable-as-Subtrahend''},
    ylabel={Accuracy},
    x label style={at={(axis description cs:0.5,-0.225)},anchor=north},
    y label style={at={(axis description cs:-0.165,0.5)},anchor=south},
    legend style={nodes={scale=0.85}, legend columns=3,anchor=north,at={(0.38,-0.57)}}
]
\addplot[
    color=NavyBlue,
    mark=o,
    line width=0.5pt,
    mark size=2.6pt,
    error bars/.cd,
    y dir=both, y explicit,
    error bar style={line width=0.7pt, color=NavyBlue},
    error mark options={rotate=90, NavyBlue, mark size=2.5pt}
    ]
    coordinates {
    (0, 0.87)
    (1, 0.84)
    (2, 0.91)
    (3, 0.80)
    (4, 0.81)
    };
    \addlegendentry{\textsc{Step-By-Step Model}}

\addplot[
    color=ForestGreen,
    mark=star,
    line width=0.5pt,
    mark size=2.6pt,
    error bars/.cd,
    y dir=both, y explicit,
    error bar style={line width=0.7pt, color=ForestGreen},
    error mark options={rotate=90, ForestGreen, mark size=2.5pt}
    ]
    coordinates {
    (0, 0.92)
    (1, 0.20)
    (2, 0.04)
    (3, 0.05)
    (4, 0.03)
    };
    \addlegendentry{\textsc{Shortcut Model}}
\end{axis}
\end{tikzpicture}
    \vspace{-1em}
   \caption{Test accuracies with increasing number of equations containing a variable as the subtrahend. The step-by-step computation model (Step-by-step Model) is evaluated on OOD 7-step problems since the accuracy of this model in both ID ones and OOD 6-step is nearly 100\%. The model trained on problems with unfixed premise order (Shortcut Model) is evaluated on ID 5-step problems.} 
    \vspace{-0.5em}
    \label{fig:comparison-stepwise-shortcut}
\end{figure}

\paragraph{Language models do not think step-by-step when the premise order is not fixed and overfit to an incorrect shortcut.}
Based on our analysis above, accuracy sharply declines if the model relies on shortcut computation. 
In contrast, step-by-step computation would result in minimal accuracy variation, as whether variables are subtrahends or not does not significantly affect sequential reasoning.
To validate our hypothesis, we plot the accuracy trends against the number of equations with ``Variable-as-Subtrahend'' in our step-by-step computation model used in Section~\ref{sec:rq1}. 
As shown in Figure~\ref{fig:comparison-stepwise-shortcut}, there is only a slight variation in the accuracy of the step-by-step computation model while the accuracy of the model trained on problems with varied premise order drops significantly, which verifies our hypothesis.\footnote{We also provide mechanistic analysis of ``Variable-as-Subtrahend'' in Appendix~\ref{appendix:shortcut-var}.}

To sum up, when the training data follows a fixed pattern, LMs can learn a fixed pattern to store each intermediate result upon completing a step.
For instance, in a forward premise order, the model simply follows the operators to compute the results of operands sequentially (i.e., step-by-step reasoning).
There is no need to track the variables, as they must come from the previous step.
However, when the premise order is shuffled, this shortcut pattern no longer exists, which necessitates the true reasoning capability: first tracking the variables and then performing the computation.
More steps involve more complex tracking and computation, which explains why accuracy decreases as the number of steps increases.
This implies that when LMs perform implicit reasoning, they are relying on shortcuts rather than engaging in true reasoning.

Furthermore, we find that the premise order does not significantly affect the model trained on an unfixed pattern. 
This further validates our hypothesis that such a language model relies on shortcuts for reasoning, as there is no difference in reasoning through shortcuts like chaining the numbers directly, whether in forward order (e.g., ``c=6+2-3+4'') or reverse order (e.g., ``c=4-3+6+2'').

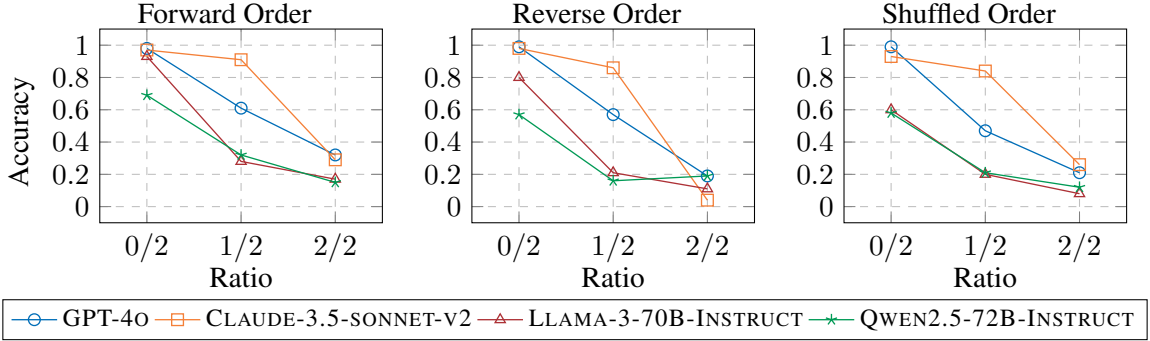
\begin{figure*}
    \centering
    \begin{minipage}{0.03\textwidth}
    \begin{tikzpicture}
        \node at (0,0) [rotate=90] {\quad\ Accuracy};
    \end{tikzpicture}
    \end{minipage}
    \begin{minipage}{0.3\textwidth}
        \centerline{Forward Order}
        \begin{tikzpicture}
        \begin{axis}[
            width=5.3cm, height=4.15cm,  
            xmin=-0.5, xmax=2.5,
            ymin=-0.1, ymax=1.1,
            xtick={0, 1, 2},
            xticklabels={$0/2$, $1/2$, $2/2$},
            ytick={0.0, 0.2, 0.4, 0.6, 0.8, 1.0},
            legend to name=sharedlegend,
            ymajorgrids=true,
            xmajorgrids=true,
            grid style=dashed,
            xlabel={Ratio},
            x label style={at={(axis description cs:0.5,0.03)},anchor=north},
            legend style={nodes={scale=0.85, legend columns=4}}
        ]
        \addplot[
            color=NavyBlue,
            mark=o,
            line width=0.5pt,
            mark size=2.2pt,
            error bars/.cd,
            y dir=both, y explicit,
            error bar style={line width=0.7pt, color=NavyBlue},
            error mark options={rotate=90, NavyBlue, mark size=2.5pt}
            ]
            coordinates {
            (0, 0.98) 
            (1, 0.61) 
            (2, 0.32) 
            };
            \addlegendentry{\textsc{GPT-4o}}

        \addplot[
            color=Orange,
            mark=square,
            line width=0.5pt,
            mark size=2.2pt,
            error bars/.cd,
            y dir=both, y explicit,
            error bar style={line width=0.7pt, color=Orange},
            error mark options={rotate=90, Orange, mark size=2.5pt}
            ]
            coordinates {
            (0, 0.97)
            (1, 0.91)
            (2, 0.29)
            };
        \addlegendentry{\textsc{Claude-3.5-sonnet-v2}}
        
        \addplot[
            color=Maroon,
            mark=triangle,
            line width=0.5pt,
            mark size=2.2pt,
            error bars/.cd,
            y dir=both, y explicit,
            error bar style={line width=0.7pt, color=Maroon},
            error mark options={rotate=90, Maroon, mark size=2.5pt}
            ]
            coordinates {
            (0, 0.93)
            (1, 0.28)
            (2, 0.17)
            };
            \addlegendentry{\textsc{Llama-3-70B-Instruct}}

        \addplot[
            color=ForestGreen,
            mark=star,
            line width=0.5pt,
            mark size=2.2pt,
            error bars/.cd,
            y dir=both, y explicit,
            error bar style={line width=0.7pt, color=ForestGreen},
            error mark options={rotate=90, ForestGreen, mark size=2.5pt}
            ]
            coordinates {
            (0, 0.69)
            (1, 0.32)
            (2, 0.15)
            };
            \addlegendentry{\textsc{Qwen2.5-72B-Instruct }}
    
        \end{axis}
        \end{tikzpicture}
    \end{minipage}
    \begin{minipage}{0.3\textwidth}
        \centerline{Reverse Order}
        \begin{tikzpicture}
        \begin{axis}[
            width=5.3cm, height=4.15cm,  
            xmin=-0.5, xmax=2.5,
            ymin=-0.1, ymax=1.1,
            xtick={0, 1, 2},
            xticklabels={$0/2$, $1/2$, $2/2$},
            ytick={0.0, 0.2, 0.4, 0.6, 0.8, 1.0},
            legend to name=sharedlegend,
            ymajorgrids=true,
            xmajorgrids=true,
            grid style=dashed,
            xlabel={Ratio},
            x label style={at={(axis description cs:0.5,0.03)},anchor=north},
            legend style={nodes={scale=0.85, legend columns=4}}
        ]
        \addplot[
            color=NavyBlue,
            mark=o,
            line width=0.5pt,
            mark size=2.2pt,
            error bars/.cd,
            y dir=both, y explicit,
            error bar style={line width=0.7pt, color=NavyBlue},
            error mark options={rotate=90, NavyBlue, mark size=2.5pt}
            ]
            coordinates {
            (0, 0.99) 
            (1, 0.57) 
            (2, 0.19) 
            };
            \addlegendentry{\textsc{GPT-4o}}

        \addplot[
            color=Orange,
            mark=square,
            line width=0.5pt,
            mark size=2.2pt,
            error bars/.cd,
            y dir=both, y explicit,
            error bar style={line width=0.7pt, color=Orange},
            error mark options={rotate=90, Orange, mark size=2.5pt}
            ]
            coordinates {
            (0, 0.98)
            (1, 0.86)
            (2, 0.04)
            };
        \addlegendentry{\textsc{Claude-3.5-sonnet-v2}}
        
        \addplot[
            color=Maroon,
            mark=triangle,
            line width=0.5pt,
            mark size=2.2pt,
            error bars/.cd,
            y dir=both, y explicit,
            error bar style={line width=0.7pt, color=Maroon},
            error mark options={rotate=90, Maroon, mark size=2.5pt}
            ]
            coordinates {
            (0, 0.80)
            (1, 0.21)
            (2, 0.11)
            };
            \addlegendentry{\textsc{Llama-3-70B-Instruct}}

        \addplot[
            color=ForestGreen,
            mark=star,
            line width=0.5pt,
            mark size=2.2pt,
            error bars/.cd,
            y dir=both, y explicit,
            error bar style={line width=0.7pt, color=ForestGreen},
            error mark options={rotate=90, ForestGreen, mark size=2.5pt}
            ]
            coordinates {
            (0, 0.57)
            (1, 0.16)
            (2, 0.19)
            };
            \addlegendentry{\textsc{Qwen2.5-72B-Instruct }}
        
        \end{axis}
        \end{tikzpicture}
    \end{minipage}
    \begin{minipage}{0.3\textwidth}
        \centerline{Shuffled Order}
        \begin{tikzpicture}
        \begin{axis}[
            width=5.3cm, height=4.15cm,  
            xmin=-0.5, xmax=2.5,
            ymin=-0.1, ymax=1.1,
            xtick={0, 1, 2},
            xticklabels={$0/2$, $1/2$, $2/2$},
            ytick={0.0, 0.2, 0.4, 0.6, 0.8, 1.0},
            legend to name=sharedlegend,
            ymajorgrids=true,
            xmajorgrids=true,
            grid style=dashed,
            xlabel={Ratio},
            x label style={at={(axis description cs:0.5,0.03)},anchor=north},
            legend style={nodes={scale=0.85}, legend columns=4}
        ]
        \addplot[
            color=NavyBlue,
            mark=o,
            line width=0.5pt,
            mark size=2.2pt,
            error bars/.cd,
            y dir=both, y explicit,
            error bar style={line width=0.7pt, color=NavyBlue},
            error mark options={rotate=90, NavyBlue, mark size=2.5pt}
            ]
            coordinates {
            (0, 0.99) 
            (1, 0.47) 
            (2, 0.21) 
            };
            \addlegendentry{\textsc{GPT-4o}}
            
        \addplot[
            color=Orange,
            mark=square,
            line width=0.5pt,
            mark size=2.2pt,
            error bars/.cd,
            y dir=both, y explicit,
            error bar style={line width=0.7pt, color=Orange},
            error mark options={rotate=90, Orange, mark size=2.5pt}
            ]
            coordinates {
            (0, 0.93)
            (1, 0.84)
            (2, 0.26)
            };
        \addlegendentry{\textsc{Claude-3.5-sonnet-v2}}
        
        \addplot[
            color=Maroon,
            mark=triangle,
            line width=0.5pt,
            mark size=2.2pt,
            error bars/.cd,
            y dir=both, y explicit,
            error bar style={line width=0.7pt, color=Maroon},
            error mark options={rotate=90, Maroon, mark size=2.5pt}
            ]
            coordinates {
            (0, 0.60)
            (1, 0.20)
            (2, 0.08)
            };
            \addlegendentry{\textsc{Llama-3-70B-Instruct}}

        \addplot[
            color=ForestGreen,
            mark=star,
            line width=0.5pt,
            mark size=2.2pt,
            error bars/.cd,
            y dir=both, y explicit,
            error bar style={line width=0.7pt, color=ForestGreen},
            error mark options={rotate=90, ForestGreen, mark size=2.5pt}
            ]
            coordinates {
            (0, 0.58)
            (1, 0.21)
            (2, 0.12)
            };
            \addlegendentry{\textsc{Qwen2.5-72B-Instruct }}
        
        \end{axis}
        \end{tikzpicture}
    \end{minipage}
    \pgfplotslegendfromname{sharedlegend}
    
    \caption{Performance comparison on 3-step problems with increasing numbers of equations containing a variable as the subtrahend. The problems in all the figures are the same, except for the order of premises. In a 3-step problem, at most two equations can have a variable as the subtrahend.}
    \vspace{-0.8em}
    \label{fig:llm-vas}
\end{figure*}

\section{How Do LLMs Perform Multi-step Implicit Reasoning?}
\label{sec:rq3}
In the previous section, we found that GPT-2 is unable to perform implicit reasoning when there is no fixed pattern to learn during training. 
Does this phenomenon also apply to current SoTA LLMs, given that their training data is not always presented in a fixed order? 
\emph{Do these models reason step-by-step, or rely on shortcuts to solve the problem?}

\paragraph{Setup.}
We conduct zero-shot experiments using both open-source and closed-source models, including GPT-4o-2024-08-06~\cite{openai2024gpt4ocard}, Claude-3.5-sonnet-20241022-v2~\cite{anthropic2024claude3}, Llama-3-70B-Instruct~\cite{llama3modelcard}, and Qwen2.5-72B-Instruct~\cite{qwen2.5}.
We instruct the model to provide answers directly with the temperature set to 0.
To ensure the consistency and fairness of our evaluation: 
1) We retain the original data generation method but restrict instances to those with intermediate computation results between 0 and 22, thus eliminating the impact of mod23 on accuracy.
2) We focus on 3-step problems, as implicit reasoning for 4-step problems proves too challenging for current LLMs, with low performance that undermines the reliability of our experiments.
3) To reduce randomness, we generate 100 problems for each ratio of equations containing a variable as the subtrahend. For each question, we evaluate it with three premise orders, i.e., forward order, reverse order, and shuffled order.
4) The accuracy is computed only in cases where the model does not output in CoT format.
More details of the experimental setups are in Appendix~\ref{appendix:prompt}.

\begin{figure}[ht]
    \centering
    \begin{tikzpicture}[scale=0.65]
    \draw[->] (0,0) -- (8.5,0) node[right] {Steps};
    \draw[->] (0,0) -- (0,5.5) node[above] {Acc.};
    
    \draw[fill=blue!40] (0.5,0) rectangle (1,4.90);
    \draw[fill=blue!40] (1.5,0) rectangle (2,4.2);
    \draw[fill=blue!40] (2.5,0) rectangle (3,4.05);
    \draw[fill=blue!40] (3.5,0) rectangle (4,3.6);
    \draw[fill=blue!40] (4.5,0) rectangle (5,3.2);
    \draw[fill=blue!40] (5.5,0) rectangle (6,2.65);
    \draw[fill=blue!40] (6.5,0) rectangle (7,2.2);
    \draw[fill=blue!40] (7.5,0) rectangle (8,1.35);
    
    \draw[dashed, red] (0,1.6) -- (8.3,1.6);
    
    \node[below] at (0.75,0) {3};
    \node[below] at (1.75,0) {4};
    \node[below] at (2.75,0) {5};
    \node[below] at (3.75,0) {6};
    \node[below] at (4.75,0) {7};
    \node[below] at (5.75,0) {8};
    \node[below] at (6.75,0) {9};
    \node[below] at (7.75,0) {10};
    
    \foreach \y/\ytext in {0/0, 1/0.2, 2/0.4, 3/0.6, 4/0.8, 5/1.0}
    {
        \draw (-0.1,\y) -- (0.1,\y);
        \node[left] at (0,\y) {\ytext};
    }
\end{tikzpicture}
    \vspace{-0.2cm}
    \caption{The accuracy of GPT-4o on problems with different step counts. The premise is in the forward order without any subtrahend being a variable. The red dashed line represents the accuracy of the same model on 3-step problems, where there are two equations with ``Variable-as-Subtrahend''.}
    \vspace{-1em}
    \label{fig:gpt4o-diff-step}
\end{figure}
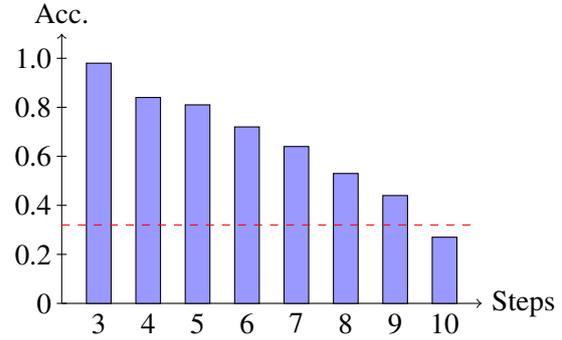

\subsection{Result}
Figure~\ref{fig:llm-vas} shows the accuracy of LLMs on problems with different ratios of equations containing a variable as the subtrahend. 
We find: 1) As the proportion of expressions with a variable as the subtrahend increases, the accuracy of the LLMs tends to decrease drastically. 
The accuracy of GPT-4o even drops from nearly 100\% to approximately 30\% regardless of premise order.
2) All the models fail to do 3-step problems containing two equations with a variable as the subtrahend, and open-source LLMs still lag behind closed-source LLMs in implicit reasoning.
3) Compared to the influence of variables as the subtrahends, the impact of premise order is not that significant, which aligns with our models trained on unfixed premise order.

We further plot the accuracy of GPT-4o on problems stated in the forward order without any subtrahend being a variable but with different steps of calculations.
From Figure~\ref{fig:gpt4o-diff-step}, we observe that though the accuracy of GPT-4o decreases gradually, the accuracy on 9-step problems even surpasses that on 3-step problems containing two equations with a variable as the subtrahend in Figure~\ref{fig:llm-vas}.

To sum up, the findings suggest that \textbf{LLMs likely rely on shortcuts for implicit reasoning rather than performing step-by-step reasoning}, which aligns with our observations in the GPT-2 model. 
To speak further, while current LLMs can perform implicit reasoning within a fixed pattern and for a limited number of steps, they cannot generalize beyond these constraints.

\section{Conclusion}
\label{sec:conclusion}
In this paper, we investigate the implicit reasoning mechanism to uncover why advanced reasoning capabilities fail to emerge in the implicit reasoning style. 
We find that language models rely on shortcuts for implicit reasoning, and these shortcuts only work when the training data aligns with a specific pattern that supports directly chaining numbers.
As a result, language models struggle with the ``Variable as Subtrahend Plight,'' which requires true reasoning capabilities, such as variable tracking and step-by-step computation, where shortcuts are no longer effective.
Experiments with current SoTA LLMs further validate our findings.

We hope this work deepens the understanding of implicit reasoning limitations in LMs and sparks future research to address LMs' key challenges in implicit multi-step reasoning.

\section*{Limitations}
\label{sec:limitation}
In this paper, we explore the mechanism of the language models performing multi-step implicit reasoning on synthetic arithmetic problems. To avoid large number operations and decimal operations, we only experiment with two fundamental arithmetic operators. This limitation suggests that future work could expand the scope to include a broader range of mathematical operators. Besides, we only focus on arithmetic reasoning due to the reasons elaborated in Section~\ref{intro}. We leave reasoning beyond arithmetic problems, e.g., commonsense reasoning, for further exploration in future research.

Another limitation lies in the possibility of inflated performance when evaluating LLMs. As their training methodologies and datasets remain proprietary, it is unclear whether these models were exposed to synthetic computational tasks similar to those explored in our study. 

\section*{Acknowledgments}
We are grateful for the constructive comments from the anonymous reviewers.
We would also like to thank Jiangjie Chen from Fudan University for valuable discussions and comments on this project.
This work is supported by the Chinese NSF Major Research Plan (No.92270121).

\bibliography{custom}

\clearpage
\appendix
\label{sec:appendix}
\section{More Details of the Data Generation Process}
\label{appendix:choice-of-k}
\begin{figure}[h]
    \centering
    \includegraphics[width=0.95\linewidth]{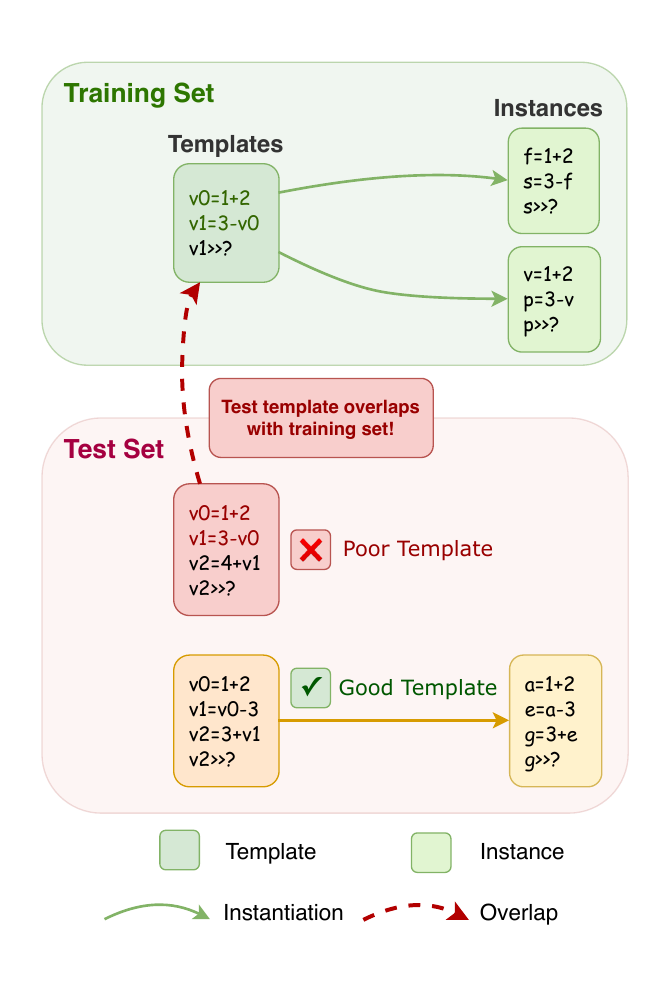}
    \caption{An overview of the data generation process.} 
    \vspace{-1em}
    \label{fig:data_gen}
\end{figure}

We provide an overview of the data generation process in Figure~\ref{fig:data_gen}. 
First, for the training set, we create \num{25000} distinct multi-step calculation templates for questions of each length (2 to 5 steps).
For the 1-step template, we include all the possible combinations to enable the model to learn basic calculations.
Then, we use the same method for the test set, but an additional filter mechanism is employed to prevent LMs from utilizing intermediate results from the training set.
As shown in Figure~\ref{fig:data_gen}, the model may directly utilize the result of $v_1$ from the training data to calculate $v_2$ by just calculating $v_2=4+v_1$.
Therefore, the test set retains only the templates whose preceding calculations, apart from the first step, do not overlap with those of the training set.  
This setting prevents LMs from memorizing intermediate results during training and recall them during testing rather than performing actual reasoning.

During generation, each of the two operators has a 50\% probability, and the variable has a 50\% probability of appearing before or after the operator (except for the first step). This results in 25\% of the steps having the variable as the subtrahend in the original training and test set. 

Since the variables in the template are sorted as $v_0$, $v_1$, ..., to prevent the model from learning the calculation order through the indices, we randomly replace them with the letters a-z.
We use $K$ different groups of variable names to instantiate each template in the training set.
For the choice of $K$, please refer to the subsequent subsection.

\subsection{Effect of the Number of Template Instantiations}
In our early experiments, we find that when \(K\) equals $1$, the model trained from scratch struggles to generalize effectively to problems outside the training set, even when these problems share the same template but have different variable names.
We attempt to adjust the training hyperparameters, including the learning rate and weight decay; however, the situation remained unchanged.
After increasing \( K \) to $2$, the model successfully handles problems with the same template but different variable names, as well as those in the test set.
So, we continue to use $K = 2$ in our experiment to ensure that the failure of generalization is caused by the model rather than our data.

\begin{figure*}[h]
    \centering
    \begin{subfigure}[b]{0.34\linewidth}
        \centering
        \includegraphics[width=5.0cm, keepaspectratio]{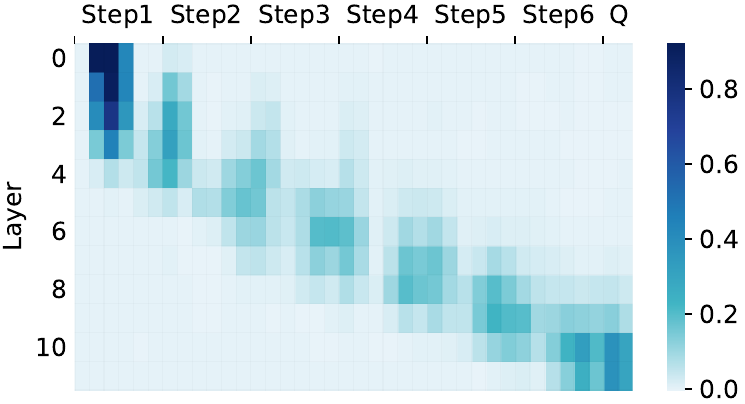}
        \caption{Logit of $r$}
        \label{fig:metric-gtcl}
    \end{subfigure}
    \hfill
    \begin{subfigure}[b]{0.33\linewidth}
        \centering
        \includegraphics[width=5.0cm, keepaspectratio]{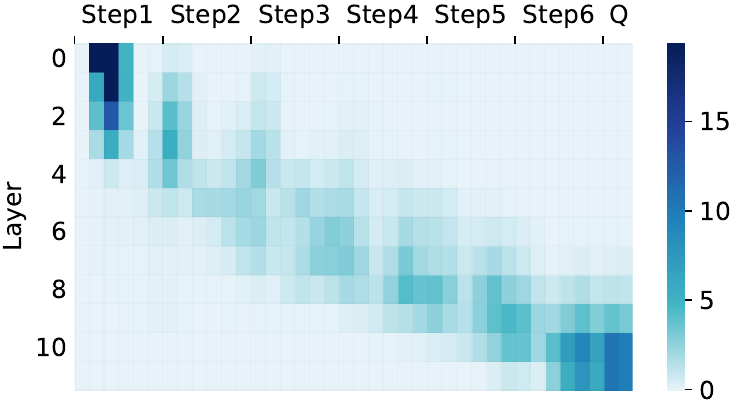}
        \caption{Logit of $r'$}
        \label{fig:metric-gtstar}
    \end{subfigure}
    \hfill
    \begin{subfigure}[b]{0.31\linewidth}
        \centering
        \includegraphics[width=5.0cm, keepaspectratio]{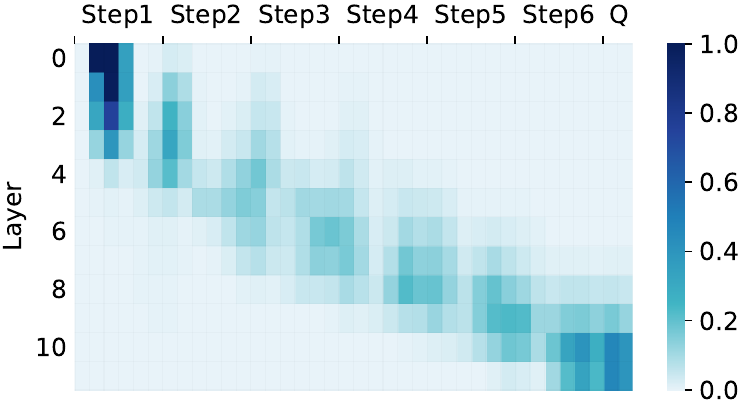}
        \caption{Logit difference}
        \label{fig:metric-logitdiff}
    \end{subfigure}
    \caption{Comparison of different patching metrics. (a) Logit of the clean run's ground truth token $r$. (b) Logit of the corrupted run's ground truth token $r'$. (c) Logit difference between $r$ and $r'$.}
    \label{fig:metric}
\end{figure*}

\begin{figure*}[h]
    \centering
    \begin{subfigure}[b]{0.45\linewidth}
        \centering
        \includegraphics[width=6.3cm, keepaspectratio]{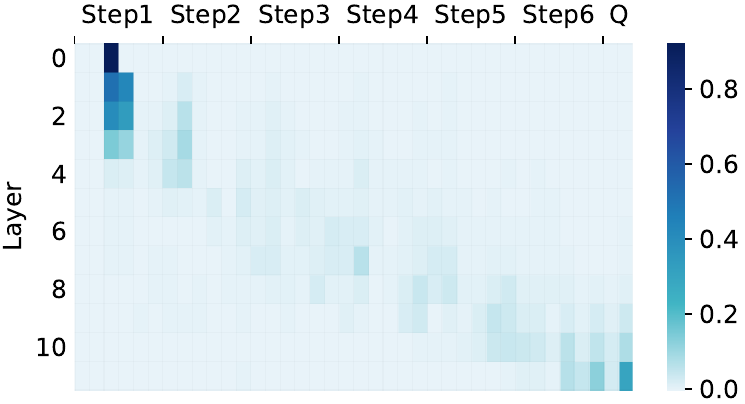}
        \caption{$1\times1$}
        \label{fig:1x1}
    \end{subfigure}
    \begin{subfigure}[b]{0.45\linewidth}
        \centering
        \includegraphics[width=6.3cm, keepaspectratio]{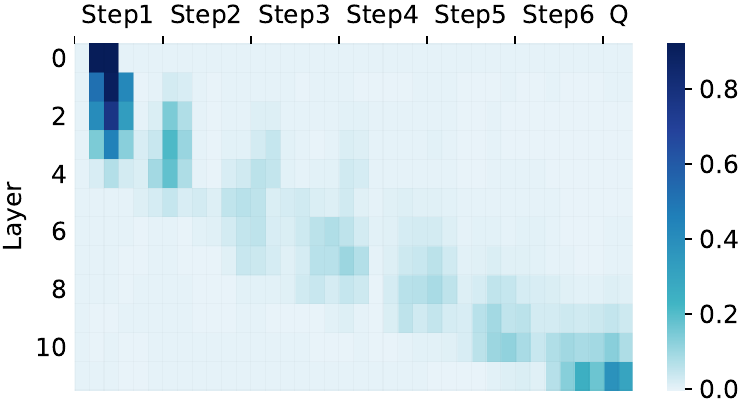}
        \caption{$1\times2$}
    \end{subfigure}
    \hfill
    \begin{subfigure}[b]{0.45\linewidth}
        \centering
        \includegraphics[width=6.3cm, keepaspectratio]{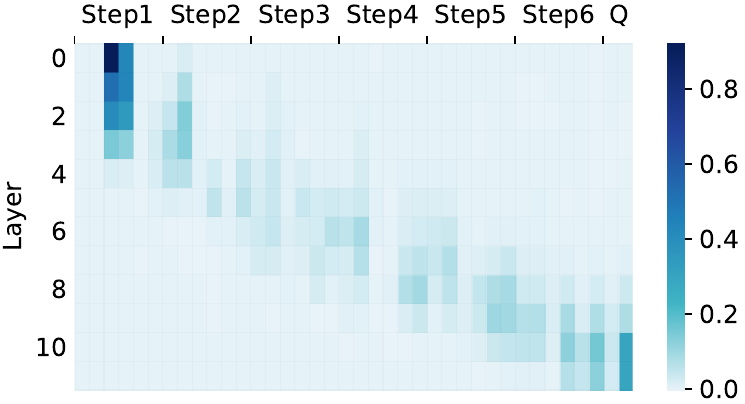}
        \caption{$2\times1$}
    \end{subfigure}
    \begin{subfigure}[b]{0.45\linewidth}
        \centering
        \includegraphics[width=6.3cm, keepaspectratio]{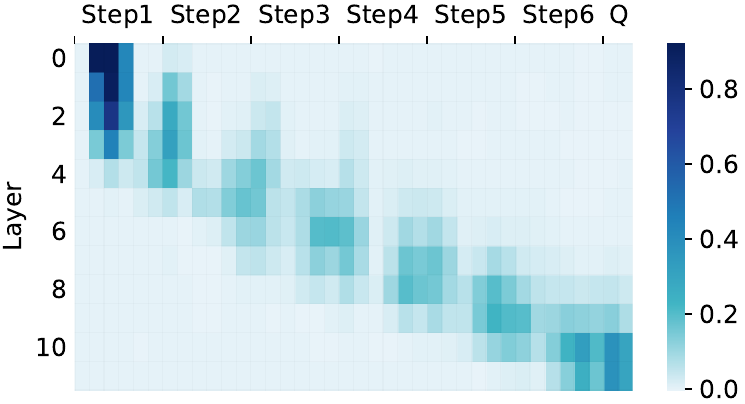}
        \caption{$2\times2$}
    \end{subfigure}
    \caption{Patching effect with different window sizes. A window size of $m\times n$ represents at each token position the representations of the region formed by the current layer and the subsequent $m-1$ layers, along with the current token and the next $n-1$ tokens, are copied from the corrupted forward pass.}
    \label{fig:window-size}
     \vspace{-1em}
\end{figure*}

\section{Choice of Activation Patching Settings}
In this section, we study the choice of metrics and window sizes in the discovery of information flow. 
\subsection{Patching Metrics}
\label{appendix:choice-of-metric}
Following the notations in Section~\ref{sec:general_setup}, $\operatorname{Logit}$ denotes the output logit at the last token position, $r$ and $r'$ are the correct answer of the original input and corrupted input, and cl, *, pt denote the clean, corrupted and patched run separately.

In Figure~\ref{fig:metric}, we compare the effect of several commonly used metrics:
\begin{itemize}[noitemsep,nolistsep,leftmargin=*]
    \item[a)] Logit of the clean run's ground-truth token $r$: $\operatorname{Logit}_\mathrm{cl}(r) - \operatorname{Logit}_\mathrm{pt}(r)$. We normalize this by $\operatorname{Logit}_\mathrm{cl}(r)$, and obtain the normalized patching effect as shown in Equation~\ref{eq:pe};
    \item[b)] Logit of the corrupted run's ground truth token $r'$: $\operatorname{Logit}_\mathrm{pt}(r') - \operatorname{Logit}_\mathrm{cl}(r')$. We do not normalize since $\operatorname{Logit}_\mathrm{cl}(r')$ can be very small, which may produce noisy localization outcomes. So we use 
    \begin{equation}
        \operatorname{PE}=\operatorname{Logit}_\mathrm{pt}(r') - \operatorname{Logit}_\mathrm{cl}(r');
    \end{equation}
    \item[c)] Logit difference: $\operatorname{LD}(r, r') = \operatorname{Logit}(r) - \operatorname{Logit}(r')$.
    We normalize this by $\operatorname{LD}_\mathrm{cl}(r, r')-\operatorname{LD}_\mathrm{*}(r, r')$, and get
    \begin{equation}
        \operatorname{PE}=\frac{\operatorname{LD}_\mathrm{cl}(r, r')-\operatorname{LD}_\mathrm{pt}(r, r')}{\operatorname{LD}_\mathrm{cl}(r, r')-\operatorname{LD}_\mathrm{*}(r, r')},
    \end{equation}
    so it typically lies in $[0, 1]$.
\end{itemize}
We find there is no significant difference in the discovery of information flow, and we use a) in our experiments for the following reasons:
\begin{itemize}[noitemsep,nolistsep,leftmargin=*]
\item[1)] Compared to c), a) can measure the patching effect when the ground truth tokens of the clean run and the corrupted run are the same.
\item[2)] Compared to b), the patching effect of a) can be normalized to $[0, 1]$ more stably.
\end{itemize}

\subsection{Window Sizes}
\label{appendix:window-size}
Following the best practices of activation patching~\citep{DBLP:conf/iclr/ZhangN24}, we initially employ single-layer interventions to identify crucial model components. 
However, as illustrated in Figure~\ref{fig:1x1}, individual layer modifications produce only marginal effects, making it difficult to isolate critical hidden states.
We speculate that language models may use aggregations from multiple inference pathways~\cite{mcgrath2023hydraeffectemergentselfrepair}, using a region rather than a hidden state to perform computations and restore intermediate results.
Noting that the critical blocks in Figure~\ref{fig:1x1} frequently exhibit rectangular patterns, we implement a $2\times2$ window size to capture the joint effect of these regions. 
Our comparison of different patching window sizes in Figure~\ref{fig:window-size} reveals that different window sizes generally preserve similar information flow characteristics, and our $2\times2$ configuration best captures the information flow. 
We do not use a larger window size since the $2\times2$ window size is enough, and a larger window size may result in inflated localization plots.

\begin{figure}[h]
    \centering
    \includegraphics[width=0.94\linewidth]{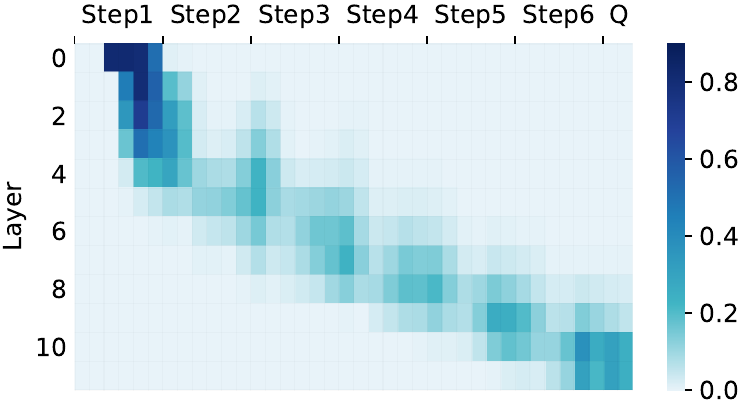}
    \caption{Activation patching on hidden states across layers and token positions when changing the first operator in the problems.} 
    \vspace{-1em}
    \label{fig:info-flow-of-first-op}
\end{figure}

\section{Information Flow Related to Operators}
We present the residual stream patching plot altering the first operator in Figure~\ref{fig:info-flow-of-first-op}. 
Similar to changing the operand, the patching effect is still pronounced at the end of each step, with information still propagating downward along the diagonal.
\label{appendix:operator-related-info-flow}

\section{Implementation Details of Masking Information from Previous Steps}
\label{appendix:attn_window_size}
In this experiment, we modify the Transformer model to restrict the attention so that each token can only attend to itself and the preceding $window\_size-1$ tokens. 
This is achieved by applying a sliding window mask, as illustrated in Algorithm~\ref{algo:mask}.
Specifically, we create an attention mask, where the parts we want to focus on are set to $0$, and the remaining positions are set to $-\infty$. 
The attention mask is then added to the attention score and passed through the softmax function. 
This procedure ensures the positions outside of the attention window are assigned zero attention after the softmax operation, effectively preventing the model from considering information from these positions.
Through this modification, we can further confirm whether the model only utilizes the information of the current step and the intermediate result from the previous step.

\begin{algorithm}
\caption{Creating Sliding Window Mask}
\label{algo:mask}
\begin{algorithmic}[1]
\Require $seq\_length$, $window\_size$
\Ensure $mask$
\State Initialize $mask$ as a $seq\_length \times seq\_length$ matrix
\For{$i \gets 0$ to $seq\_length - 1$}
    \For{$j \gets 0$ to $seq\_length - 1$}
        \If{$j < \max(0, i - window\_size + 1)$ \textbf{or} $j > i$}
            \State $mask[i][j] \gets -\infty$
        \Else
            \State $mask[i][j] \gets 0$
        \EndIf
    \EndFor
\EndFor
\State \Return $mask$
\end{algorithmic}
\end{algorithm}

\section{More Details of the Training Premise Pattern}
\label{appendix:premise_pattern}
\begin{figure}[h]
\includegraphics[width=\linewidth]{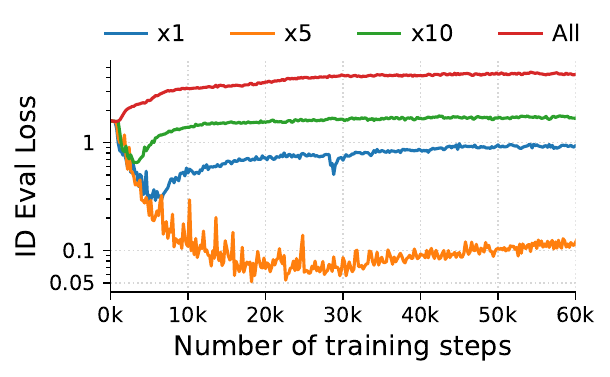}
    \caption{Loss of the model on the ID test set during training when trained with different data configurations. The y-axis is plotted on a logarithmic scale.} 
    \vspace{-0.5em}
    \label{fig:ablation_data_pattern}
\end{figure}

In order to enable the model to learn to reason when the order of the premises is not fixed, the training data needs to contain patterns with different premise orders. 
To this end, in our early experiment, we have tried several data configurations on original GPT-2-Medium to select the best one, and see whether the failure stems from the insufficiency of premise pattern.
Specifically, we gradually expand the dataset by controlling the upper limit of the patterns that can be added to the training data for each template ($\times m$ indicates that at most $m$ orders for each template will be added to the training data). 
If the total number of premise orders for a certain template is less than or equal to $m$, then all order combinations will be added to the dataset; if the total number of orders is greater than $m$, then for each template, $m$ randomly selected orders of this template will be added to the dataset. 
We provide the sample size for each dataset in Table~\ref{tab:stat-data-pattern}. 
As shown in Figure~\ref{fig:ablation_data_pattern}, including more patterns does not necessarily improve performance. When all the patterns are added, the loss of the model on the test set basically cannot decrease, and overfitting occurs rapidly. 

\begin{table}[h]
\centering
\small
\begin{tabular}{ccccc}
\toprule
\textbf{Dataset} & \textbf{$\times$1} & \textbf{$\times$5} & \textbf{$\times$10} & \textbf{All} \\
\midrule
Size & 202K & 850K & 1.4M & 7.6M \\
\bottomrule
\end{tabular}
\caption{Size of different datasets.  We control the upper limit of patterns that can be added to the dataset for each template. For example, $\times 5$ indicates that at most $5$ orders for each template will be added to the training data. }
\vspace{-2em}
\label{tab:stat-data-pattern}
\end{table}

For the dataset including all the premise orders for every template, we also try another approach that upsample problems with fewer than 5 steps to the same proportion as that in the $\times$5 dataset. Although the minimum value of the evaluation loss is slightly lower, the loss increases rapidly after reaching the minimum value.
We provide the accuracy of both models in Table~\ref{tab:ablation-data-pattern}. We find that both models cannot escape ``Variable as Subtrahend Plight''. 
Considering training and data efficiency, unless otherwise specified, we set the upper limit of the number of patterns for each step to 5 in all of our experiments. 

\begin{table}[h]
    \centering
    \small
    \begin{tabular}{lccccc}
    \toprule
    \multirow{2.5}{*}{\textbf{Order}} & \multicolumn{5}{c}{\resizebox{0.58\linewidth}{!}{\textbf{\textsc{\#Variable Being Subtrahend}}}}
    \\ \cmidrule(lr){2-6} 
         & $\mathbf{0/4}$  & $\mathbf{1/4}$ & $\mathbf{2/4}$ & $\mathbf{3/4}$ & $\mathbf{4/4}$  \\ 
        \midrule
\rowcolor[gray]{0.95} \multicolumn{6}{c}{\textit{$\times 5$}} \\ \midrule
        Forward & 1.00 & 0.90 & 0.90 & 0.30 & 0.05  \\ 
        Reverse & 1.00 & 0.97 & 0.98 & 0.50 & 0.08  \\
        Random &  1.00 & 0.83 & 0.89 & 0.45 & 0.23 \\ 
\midrule
\rowcolor[gray]{0.95} \multicolumn{6}{c}{\textit{All (Upsample)}} \\ \midrule
        Forward & 1.00 & 0.87 & 0.85 & 0.55 & 0.08  \\ 
        Reverse & 1.00 & 0.97 & 0.95 & 0.64 & 0.06  \\
        Random &  1.00 & 0.86 & 0.92 & 0.54 & 0.16  \\ 
 \bottomrule
    \end{tabular}
\caption{Accuracy of the models with different training dataset on 5-step problems.}
\label{tab:ablation-data-pattern}
\end{table}

\section{Extended Experiments on Increased Data Volume and Different Models}
\label{appendix:result-of-other-model}
\subsection{Model Size and Initialization}
\label{appendix:init}
When training from scratch, we also test larger models, i.e., GPT2-RoPE-medium, by increasing the number of layers from 12 to 24, but the accuracy does not improve. 
We find that our GPT2-RoPE model initiated from a pre-trained GPT-2's weight may alleviate overfitting and have a higher performance, but we only observe this phenomenon on GPT2-RoPE-Medium.
As shown in Table~\ref{tab:gpt2-rope-pretrained-acc}, despite the increased accuracy, the model still fails when almost all the variables are subtrahends.
In addition, investigating model initialization is not the main focus of our paper.
\begin{table}[h]
    \centering
    \small
    \begin{tabular}{lccccc}
    \toprule
    \multirow{2.5}{*}{\textbf{Order}} & \multicolumn{5}{c}{\resizebox{0.58\linewidth}{!}{\textbf{\textsc{\#Variable Being Subtrahend}}}}
    \\ \cmidrule(lr){2-6} 
         & $\mathbf{0/4}$  & $\mathbf{1/4}$ & $\mathbf{2/4}$ & $\mathbf{3/4}$ & $\mathbf{4/4}$  \\ 
        \midrule
\rowcolor[gray]{0.95} \multicolumn{6}{c}{\textit{GPT2-RoPE-Medium-Pretrained}} \\ \midrule
        Forward & 1.00 & 0.98 & 0.99 & 0.97 & 0.05  \\ 
        Reverse & 1.00 & 0.99 & 0.84 & 0.06 & 0.02 \\
        Random & 1.00 & 0.90 & 0.83 & 0.35 & 0.08 \\ 
 \bottomrule
    \end{tabular}
\caption{Accuracy of the model on 5-step questions with different numbers of variables being subtrahends.}
\label{tab:gpt2-rope-pretrained-acc}
\end{table}

\begin{figure}[h]
\includegraphics[width=\linewidth]{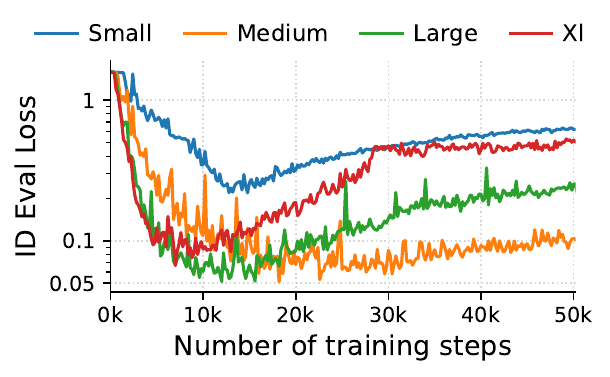}
    \caption{Loss of the model on the ID test set during training when trained with different model sizes. The y-axis is plotted on a logarithmic scale.} 
    \label{fig:ablation_model_size}
\end{figure}

Since training from a pre-trained model is often better than training from scratch, in order to better control the experimental variables and illustrate the impact of the model size on the experimental results, we use the original pre-trained GPT-2 series to explore the influence of the model size. In Figure~\ref{fig:ablation_model_size}, we present the loss curves on the ID test set for different model sizes, ranging from GPT2-Small (124M) to GPT2-XL (1.5B). We find that increasing the model size from Small to Medium can lead to improvements. However, after reaching a certain size (Medium), further increasing the model size does not yield additional gains.

\subsection{Model Architecture}
In addition to the GPT-2 models (with or without RoPE), we also test other model architecture such as Qwen2.5. 
Since the performance of the pre-trained models is better than those trained from scratch, we initiate from the pre-trained weight.
As shown in Table~\ref{tab:qwen-acc}, we find that the model still does not escape ``Variable as Subtrahend Plight''.
\begin{table}[h]
    \centering
    \small
    \begin{tabular}{lccccc}
    \toprule
    \multirow{2.5}{*}{\textbf{Order}} & \multicolumn{5}{c}{\resizebox{0.58\linewidth}{!}{\textbf{\textsc{\#Variable Being Subtrahend}}}}
    \\ \cmidrule(lr){2-6} 
         & $\mathbf{0/4}$  & $\mathbf{1/4}$ & $\mathbf{2/4}$ & $\mathbf{3/4}$ & $\mathbf{4/4}$  \\ 
        \midrule
\rowcolor[gray]{0.95} \multicolumn{6}{c}{\textit{Qwen2.5-1.5B-Base}} \\ \midrule
        Forward & 1.00 & 0.98 & 0.79 & 0.62 & 0.40   \\ 
        Reverse & 0.99 & 0.97 & 0.98 & 0.90 & 0.69  \\
        Random & 1.00 & 0.89 & 0.81 & 0.74 & 0.61   \\ 
 \bottomrule
    \end{tabular}
\caption{Accuracy of the models with different architecture on 5-step problems.}
\vspace{-1em}
\label{tab:qwen-acc}
\end{table}

\subsection{Data Volume}
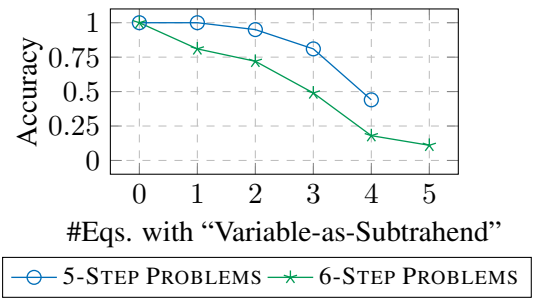
\begin{figure}[h]
    \centering
    \pgfplotsset{width=0.8\linewidth,height=0.49\linewidth,compat=1.18}
\begin{tikzpicture}
\begin{axis}[
    xmin=-0.5, xmax=5.5,
    ymin=-0.1, ymax=1.1,
    xtick={0, 1, 2, 3, 4, 5, 6},
    ytick={0.0, 0.25, 0.5, 0.75, 1.0},
    ymajorgrids=true,
    xmajorgrids=true,
    grid style=dashed,
    xlabel={\#Eqs. with ``Variable-as-Subtrahend''},
    ylabel={Accuracy},
    x label style={at={(axis description cs:0.5,-0.225)},anchor=north},
    y label style={at={(axis description cs:-0.165,0.5)},anchor=south},
    legend style={nodes={scale=0.85}, legend columns=3,anchor=north,at={(0.45,-0.5)}}
]
\addplot[
    color=NavyBlue,
    mark=o,
    line width=0.5pt,
    mark size=2.6pt,
    error bars/.cd,
    y dir=both, y explicit,
    error bar style={line width=0.7pt, color=NavyBlue},
    error mark options={rotate=90, NavyBlue, mark size=2.5pt}
    ]
    coordinates {
    (0, 1.00)
    (1, 1.00)
    (2, 0.95)
    (3, 0.81)
    (4, 0.44)
    };
    \addlegendentry{\textsc{5-Step Problems}}

\addplot[
    color=ForestGreen,
    mark=star,
    line width=0.5pt,
    mark size=2.6pt,
    error bars/.cd,
    y dir=both, y explicit,
    error bar style={line width=0.7pt, color=ForestGreen},
    error mark options={rotate=90, ForestGreen, mark size=2.5pt}
    ]
    coordinates {
    (0, 1.00)
    (1, 0.81)
    (2, 0.72)
    (3, 0.49)
    (4, 0.18)
    (5, 0.11)
    };
    \addlegendentry{\textsc{6-Step Problems}}
\end{axis}
\end{tikzpicture}
   \caption{Test accuracies with increasing number of equations containing a variable as the subtrahend. To test genuine reasoning abilities, the premise order used during testing is random.} 
    \vspace{-0.5em}
    \label{fig:ablation-data-volume-6step}
\end{figure}

Another potential explanation for the poor generalization could be the limited number of templates used during training. 
To investigate, we expand the dataset to include 50000 different templates for each step (except for the single-step question). 
Since the performance of the GPT2-RoPE model used in Table~\ref{tab:gpt2-rope-pretrained-acc} is better than the one used in Table~\ref{tab:rq2-acc-diff-step}, we continue to train from the GPT2-RoPE-Medium model initiated from a pre-trained GPT-2's weight in this experiment.
As shown in Table~\ref{tab:ablation-data-volume}, this expanded experiment yields similar results, with models still failing to generalize to problems in which most variables are subtrahends.
Furthermore, we scale up the templates for 5-step problems tenfold (simultaneously upsample instances of other step counts to maintain the proportion of different step counts within the dataset), so the training dataset comprises 500K different 5-step templates, hoping the model will thoroughly learn to solve 5-step problems.
In addition to the accuracy on 5-step problems in Table~\ref{tab:ablation-data-volume}, we also visualize the model's performance on 6-step problems in Figure~\ref{fig:ablation-data-volume-6step}.
Although continuing to scale up the data can slightly boost the model’s performance within the sequence lengths seen during training, data scaling does not address the core reasoning flaw that the model cannot genuinely track variables and perform step-by-step calculations, causing it to fall into the ``Variable as Subtrahend Plight''.

\begin{table}[h]
    \centering
    \small
    \begin{tabular}{lccccc}
    \toprule
    \multirow{2.5}{*}{\textbf{Order}} & \multicolumn{5}{c}{\resizebox{0.58\linewidth}{!}{\textbf{\textsc{\#Variable Being Subtrahend}}}}
    \\ \cmidrule(lr){2-6} 
         & $\mathbf{0/4}$  & $\mathbf{1/4}$ & $\mathbf{2/4}$ & $\mathbf{3/4}$ & $\mathbf{4/4}$  \\ 
        \midrule
\rowcolor[gray]{0.95} \multicolumn{6}{c}{\textit{50000 Templates}} \\ \midrule
        Forward & 1.00 & 0.99 & 0.97 & 0.89 & 0.25 \\
        Reverse & 1.00 & 1.00 & 0.96 & 0.85 & 0.24 \\
        Random  & 1.00 & 0.97 & 0.92 & 0.70 & 0.26 \\
\midrule
\rowcolor[gray]{0.95} \multicolumn{6}{c}{\textit{50000 Templates w/ further expansion}} \\ \midrule
        Forward & 1.00 & 1.00 & 1.00 & 0.92 & 0.47  \\ 
        Reverse & 1.00 & 1.00 & 0.98 & 0.90 & 0.44  \\
        Random &  1.00 & 1.00 & 0.95 & 0.81 & 0.44  \\ 
 \bottomrule
    \end{tabular}
\caption{Accuracy of the models with different training data volume on 5-step problems.}
\vspace{-1em}
\label{tab:ablation-data-volume}
\end{table}

\section{Mechanistic Insights into ``Variable as Subtrahend Plight''}
\label{appendix:shortcut-var}
Due to the low accuracy of the models trained from scratch, we use the GPT2-RoPE-Medium-Pretrained model from Section~\ref{appendix:init} instead for analysis.
Since the model can not fully learn to do implicit reasoning on problems requiring more than 3 steps of reasoning, we first restrict our analysis to 3-step problems. 
We use $1\times 1$ patching for this model, since $1\times 1$ patching has already had a noticeable impact.

\begin{figure*}[t]
    \centering
    \begin{subfigure}[b]{0.45\linewidth}
        \centering
        \includegraphics[width=6.3cm, keepaspectratio]{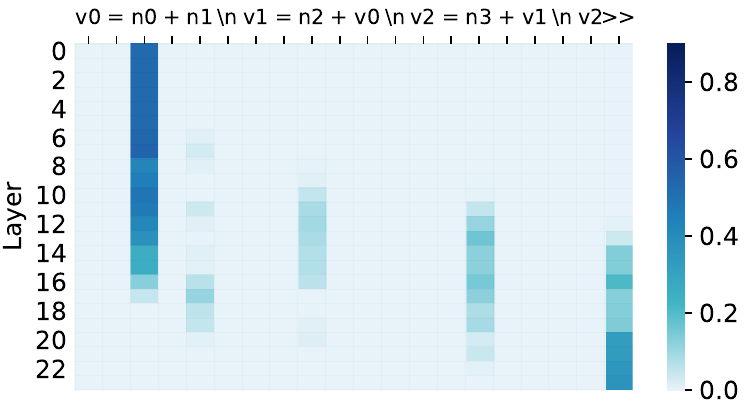}
        \caption{$\operatorname{number}+\operatorname{variable}$}
    \end{subfigure}
    \begin{subfigure}[b]{0.45\linewidth}
        \centering
        \includegraphics[width=6.3cm, keepaspectratio]{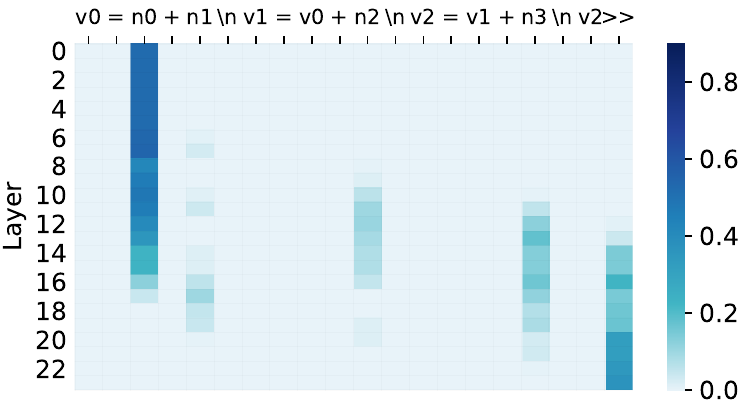}
        \caption{$\operatorname{variable}+\operatorname{number}$}
    \end{subfigure}
    \hfill
    \begin{subfigure}[b]{0.45\linewidth}
        \centering
        \includegraphics[width=6.3cm, keepaspectratio]{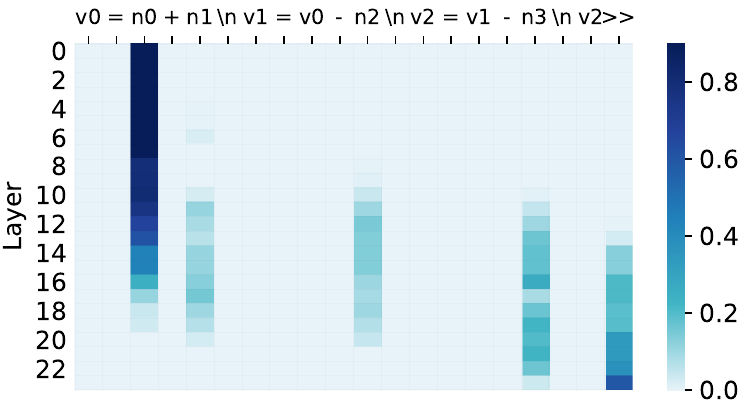}
        \caption{$\operatorname{variable}-\operatorname{number}$}
    \end{subfigure}
    \begin{subfigure}[b]{0.45\linewidth}
        \centering
        \includegraphics[width=6.3cm, keepaspectratio]{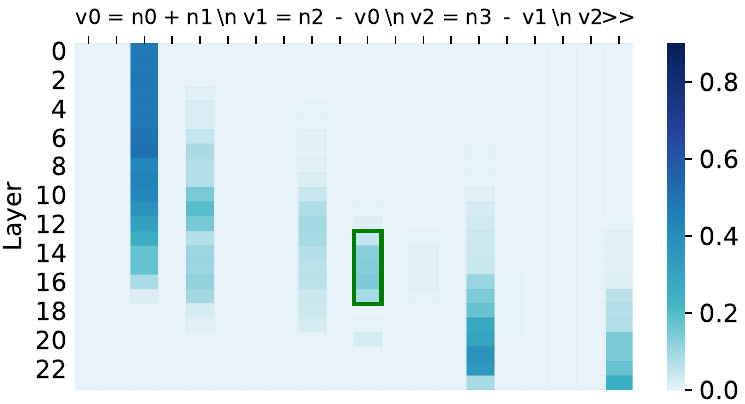}
        \caption{$\operatorname{number}-\operatorname{variable}$}
        \label{fig:step3-123-num-var}
    \end{subfigure}
    \caption{Patching effect with different combination of the operator and the position of the variable when changing the first number in the problem.}
    \label{fig:pe-diff-pos}
\end{figure*}

\begin{figure*}[t]
    \centering
    \begin{subfigure}[b]{0.32\linewidth}
        \centering
        \includegraphics[width=5.2cm, keepaspectratio]{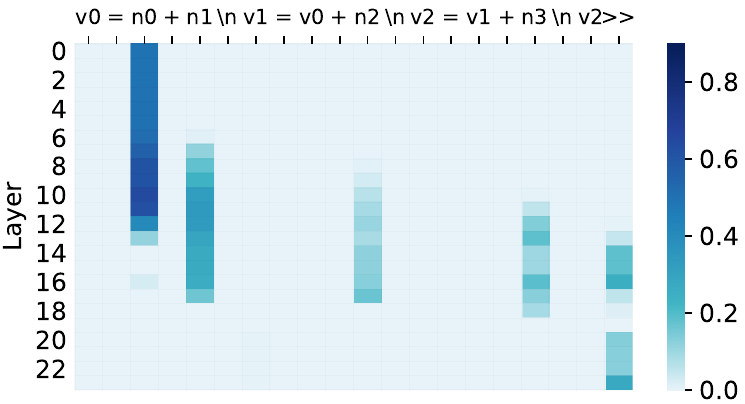}
        \caption{Forward order}
    \end{subfigure}
    \hfill
    \begin{subfigure}[b]{0.32\linewidth}
        \centering
        \includegraphics[width=5.2cm, keepaspectratio]{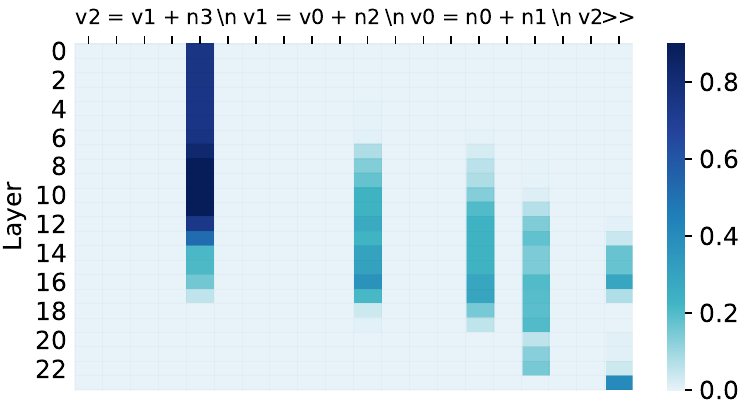}
        \caption{Reverse order}
    \end{subfigure}
    \hfill
    \begin{subfigure}[b]{0.32\linewidth}
        \centering
        \includegraphics[width=5.2cm, keepaspectratio]{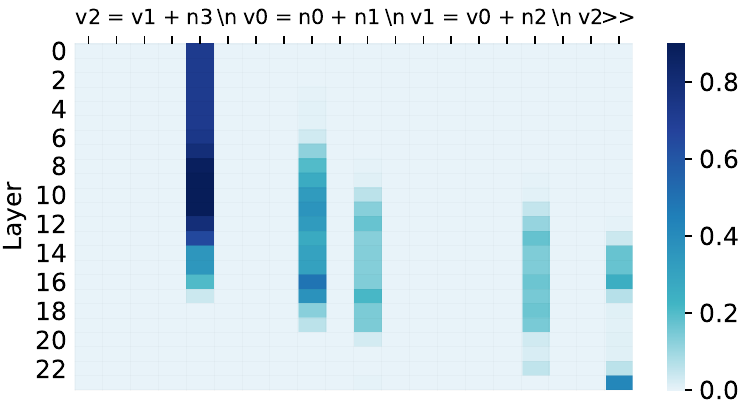}
        \caption{Shuffled order}
    \end{subfigure}
    \caption{Patching effect of different premise order averaged on the same set of problems when changing the first number in the problem.}
    \label{fig:pe-diff-order}
\end{figure*}

\begin{figure*}[t]
    \centering
    \begin{subfigure}[b]{0.45\linewidth}
        \centering
        \includegraphics[width=6.3cm, keepaspectratio]{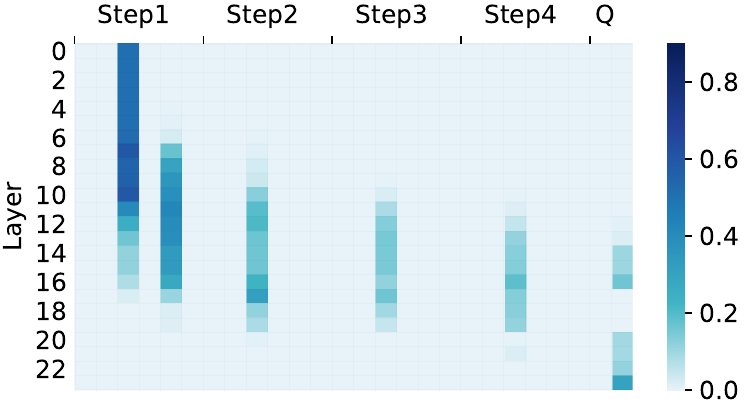}
        \caption{No variables in the problem are subtrahends}
    \end{subfigure}
    \begin{subfigure}[b]{0.45\linewidth}
        \centering
        \includegraphics[width=6.3cm, keepaspectratio]{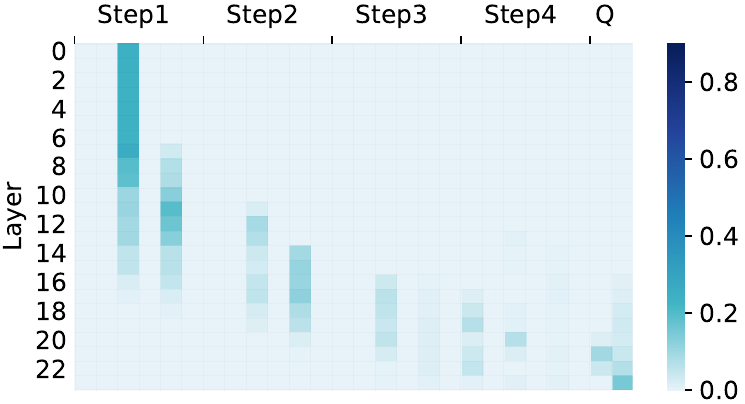}
        \caption{Only the variable in the second step is subtrahend}
    \end{subfigure}

    \caption{Patching effect on 4-step problems when changing the first number. Only the second steps of the problems are different.}
    \label{fig:pe-step2-diff}
\end{figure*}

To see why the model fails to handle equations with variables being the subtrahends, we begin our mechanistic exploration by investigating the impact of the position of the variables. 
Specifically, we analyze four distinct operator-variable combinations: ``$\operatorname{number}+\operatorname{variable}$'', ``$\operatorname{variable}+\operatorname{number}$'', ``$\operatorname{variable}-\operatorname{number}$'' and ``$\operatorname{number}-\operatorname{variable}$''.
As shown in Figure~\ref{fig:pe-diff-pos}, the first three graphs exhibit similar patterns, with the exception of the fourth graph, which shows some differences. 
We can see that in the first three graphs, the darker-colored areas are exclusively distributed in the output and numerical tokens, which means that the information in the remaining positions has no effect on the output.
This phenomenon holds for all premise orders (Figure~\ref{fig:pe-diff-order}), since premise order does not disturb implicit reasoning through chaining the numbers directly.
In contrast, in Figure~\ref{fig:step3-123-num-var}, we find that the dark color appears on the variable token (i.e., $v_0$), which means the model needs the variable value at the subtrahend position to handle subsequent calculations.
We also provide the patching plot on 4-step problems in Figure~\ref{fig:pe-step2-diff}, where a clear difference can also be observed.

These mechanistic findings show that LMs chain the numbers directly when there is no variable as the subtrahend, and explain why the premise order does not significantly affect accuracy, which validates our previous analysis in Section~\ref{sec:rq2} that the model relies on shortcuts to solve the problems.

\section{More Details of the Experimental Setup in Section~\ref{sec:rq3}}
\label{appendix:prompt}
In our preliminary tests, GPT-4o achieved less than 35\% accuracy on 4-step problems containing only one variable as the subtrahend, while other open-source models performed only slightly above random guessing.
Thus, we only study 3-step problems to ensure meaningful evaluation and better show the decreasing trend of the accuracy. 
Since the first step of the operation is between numbers, there are at most two equations containing a variable as the subtrahend in 3-step problems.

As the random premise order may still contain the forward order and the reverse order, we specify a fixed shuffled order instead.
Specifically, we rearrange the original premise ([step1, step2, step3]) to [step3, step1, step2].
The second step is delayed until the end, so the model can only link all the steps together at the last of the problem.

To prevent generic LLMs from using CoT reasoning to answer the question, we carefully craft the prompt to instruct the model to directly output the answer. An example of the prompt used for instructing generic LLMs to think without extra tokens in our task is shown below.

For Qwen and Llama, we use
\lstset{
    backgroundcolor=\color[RGB]{245,245,245},
    breaklines=true,
    breakindent=0pt,
    basicstyle=\ttfamily\small,
    frame=trbl,
    frameround = tttt,
}\begin{lstlisting}
a = 4 + 14
c = a - 12
s = 6 - c
What is the value of s? Please answer directly with "s = xx".
\end{lstlisting}
and for GPT-4o and Claude, we use
\lstset{
    backgroundcolor=\color[RGB]{245,245,245},
    breaklines=true,
    breakindent=0pt,
    basicstyle=\ttfamily\small,
    frame=trbl,
    frameround = tttt,
}\begin{lstlisting}
a = 4 + 14
c = a - 12
s = 6 - c
What is the value of s? You must answer directly. Only output the final result. Begin your answer with "s = xx".
\end{lstlisting}
to prevent the model from outputting CoT process. 

We also test questions in the form of natural language, and reach the same conclusion as shown in Table~\ref{tab:nl-acc}. This indicates that our findings are unrelated to the form of the description.

\begin{table}[h]
    \centering
    \small
    \begin{tabular}{lccccc}
    \toprule
    \multirow{2}{*}{\textbf{Order}} & \multicolumn{5}{c}{\resizebox{0.58\linewidth}{!}{\textbf{\textsc{\#Variable Being Subtrahend}}}} \\
    \cmidrule(lr){2-6} 
         & $\mathbf{0/2}$ & \hspace{1.6em} & $\mathbf{1/2}$ & \hspace{1.6em} & $\mathbf{2/2}$ \\ 
    \midrule
    \rowcolor[gray]{0.95} \multicolumn{6}{c}{\textit{GPT-4o}} \\ \midrule
        Forward & 0.94& & 0.47 && 0.28 \\ 
        Reverse & 0.93& & 0.33 && 0.21 \\
        Shuffled & 0.88& & 0.39 && 0.15 \\ 
    \midrule
    \rowcolor[gray]{0.95} \multicolumn{6}{c}{\textit{Claude-3.5-sonnet-v2}} \\ \midrule
        Forward & 0.98  & & 0.79 && 0.35 \\ 
        Reverse & 0.91  && 0.67 && 0.05 \\
        Shuffled & 0.85 && 0.64 && 0.20 \\ 
    \bottomrule
    \end{tabular}
    \caption{Performance comparison on 3-step problems in the form of natural language. The problems in each column are the same, except for the premise order.}
    \vspace{-1em}
    \label{tab:nl-acc}
\end{table}

An example of the natural language form question is shown below. Here, we convert the equations in the original prompt into natural language descriptions resembling grade school math problems.

\lstset{
    backgroundcolor=\color[RGB]{245,245,245},
    breaklines=true,
    breakindent=0pt,
    basicstyle=\ttfamily\small,
    frame=trbl,
    frameround = tttt,
}\begin{lstlisting}
A's number of apples equals 4 plus 14. 
C's number of apples equals A's number of apples minus 12. 
S's number of apples equals 6 minus C's number of apples. 
How many apples does S have? Only output the final result. Do not output intermediate results.
\end{lstlisting}

\end{document}